\documentclass[preprint,12pt]{elsarticle}

\usepackage{amssymb}

\usepackage{hyperref}
\usepackage{url}
\usepackage[cmex10]{amsmath}
\usepackage{amsfonts}
\usepackage[noend]{algorithmic}
\usepackage{array}
\usepackage{multirow}
\usepackage{caption}
\usepackage{subcaption}
\usepackage[]{algorithm2e}
\usepackage{verbatim}
\usepackage{perpage}

\usepackage{amsthm}
\usepackage{graphicx}
\usepackage{booktabs}
\usepackage{lineno}

\MakePerPage{footnote} 

\newcommand{\argmin}{\operatornamewithlimits{argmin}}

\renewcommand{\footnotesize}{\small}
\urldef{\mailsa}\path|seyedmostafa.kia@unitn.it|
\newdefinition{mydefs}{Definition}
\newdefinition{myprobs}{Problem}

\newtheorem{myprops}{Proposition}

\journal{arXiv}

\begin{document}

\begin{frontmatter}



\title{Interpretability of Multivariate Brain Maps in Brain Decoding: Definition and Quantification}
\author{Seyed Mostafa Kia\corref{cor1} \fnref{aff1,aff2,aff3}} \ead{seyedmostafa.kia@unitn.it}
\fntext[aff1]{University of Trento, Trento, Italy}
\fntext[aff2]{Fondazione Bruno Kessler (FBK), Trento, Italy}
\fntext[aff3]{Centro Interdipartimentale Mente e Cervello (CIMeC), Trento, Italy}
\cortext[cor1]{Corresponding author:}


\begin{abstract}
Brain decoding is a popular multivariate approach for hypothesis testing in neuroimaging. Linear classifiers are widely employed in the brain decoding paradigm to discriminate among experimental conditions. Then, the derived linear weights are visualized in the form of multivariate brain maps to further study the spatio-temporal patterns of underlying neural activities. It is well known that the brain maps derived from weights of linear classifiers are hard to interpret because of high correlations between predictors, low signal to noise ratios, and the high dimensionality of neuroimaging data. Therefore, improving the interpretability of brain decoding approaches is of primary interest in many neuroimaging studies. Despite extensive studies of this type, at present, there is no formal definition for interpretability of multivariate brain maps. As a consequence, there is no quantitative measure for evaluating the interpretability of different brain decoding methods. In this paper, first, we present a theoretical definition of interpretability in brain decoding; we show that the interpretability of multivariate brain maps can be decomposed into their reproducibility and representativeness. Second, as an application of the proposed theoretical definition, we formalize a heuristic method for approximating the interpretability of multivariate brain maps in a binary magnetoencephalography (MEG) decoding scenario. Third, we propose to combine the approximated interpretability and the performance of the brain decoding model into a new multi-objective criterion for model selection. Our results for the MEG data show that optimizing the hyper-parameters of the regularized linear classifier based on the proposed criterion results in  more informative multivariate brain maps. More importantly, the presented definition provides the theoretical background for quantitative evaluation of interpretability, and hence, facilitates the development of more effective brain decoding algorithms in the future.
\end{abstract}


\end{frontmatter}


\section{Introduction}
\label{sec:introduction}
Understanding the mechanisms of the brain has been a crucial topic throughout the history of science. Ancient Greek philosophers envisaged different functionalities for the brain ranging from cooling the body to acting as the seat of the rational soul and the center of sensation~\cite{crivellato2007soul}. Modern cognitive science, emerging in the 20th century, provides better insight into the brain's functionality. In cognitive science, researchers usually analyze recorded brain activity and behavioral parameters to discover the answers of \emph{where}, \emph{when}, and \emph{how} a brain region participates in a particular cognitive process.

To answer the key questions in cognitive science, scientists often employ mass-univariate hypothesis testing methods to test scientific hypotheses on a large set of independent variables~\cite{groppe2011amass,maris2007nonparametric}. Mass-univariate hypothesis testing is based on performing multiple tests, e.g., t-tests, one for each unit of the neuroimaging data, i.e., independent variables. Although the high spatial and temporal granularity of the univariate tests provides good interpretability of results, the high dimensionality of neuroimaging data requires a large number of tests, which reduces the sensitivity of these methods after multiple comparison correction. Although some techniques such as the non-parametric cluster-based permutation test~\cite{maris2007nonparametric} provide more sensitivity because of the cluster assumption, they still experience low sensitivity to brain activities that are narrowly distributed in time and space~\cite{groppe2011amass,groppe2011bmass}. The multivariate counterparts of mass-univariate analysis, known generally as multivariate pattern analysis (MVPA), have the potential to overcome these deficits. Multivariate approaches are capable of identifying complex spatio-temporal interactions between different brain areas with higher sensitivity and specificity than univariate analysis~\cite{van2009interpreting}, especially in group analysis of neuroimaging data~\cite{davis2014differences}.

\emph{Brain decoding}~\cite{haynes2006decoding} is an MVPA technique that provides a model based on the recorded brain signal to predict the mental state of a human subject. There are two potential applications for brain decoding: 1) brain-computer interfaces (BCIs)~\cite{wolpaw2002brain,waldert2008hand,van2009attention,nicolas2012brain}, and 2) multivariate hypothesis testing~\cite{bzdok2016classical}. In the first case, a brain decoder with maximum prediction power is desired. In the second case, in addition to the prediction power, extra information on the spatio-temporal nature of a cognitive process is desired. In this study, we are interested in the second application of brain decoding, which can be considered a multivariate alternative for mass-univariate hypothesis testing.

In brain decoding, generally, linear classifiers are used to assess the relation between independent variables, i.e., features, and dependent variables, i.e., cognitive tasks~\cite{pereira2008machine,lemm2011introduction,besserve2007classification}. This assessment is performed by solving a linear optimization problem that assigns weights to each independent variable. Currently, brain decoding is the gold standard in multivariate analysis for functional magnetic resonance imaging (fMRI)~\cite{haxby2001distributed,cox2003functional,mitchell2004learning,norman2006beyond} and magnetoencephalogram/electroencephalogram (MEEG) studies~\cite{Parra2003Singletrial,rieger2008predicting,carroll2009prediction,chan2011decoding,huttunen2013mind,
vidaurre2013survey,abadi2015affective}. It has been shown that brain decoding can be used in combination with brain encoding~\cite{naselaris2011encoding} to infer the causal relationship between stimuli and responses~\cite{weichwald2015causal}.

\emph{Brain mapping}~\cite{kriegeskorte2006information} is a higher form of neuroimaging that assigns pre-computed quantities, e.g., univariate statistics or weights of a linear classifier, to the spatio-temporal representation of neuroimaging data. In MVPA, brain mapping uses the learned parameters from brain decoding to produce brain maps, in which the engagement of different brain areas in a cognitive task is visualized. The interpretability of a brain decoder generally refers to the level of information that can be reliably derived by an expert from the resulting maps. From the neuroscientific perspective, a brain map is considered \emph{interpretable} if it enables the scientist to answer \emph{where}, \emph{when}, and \emph{how} questions. 

Typically, a trained classifier provides a black box that predicts the label of an unseen data point with some accuracy. \citet{valverde2014100} experimentally showed that in a classification task optimizing only classification error rate is insufficient to capture the transfer of crucial information from the input to the output of a classifier. It is also shown by \citet{ramdas2016classification} that in the case of data with small sample size, high dimensionality, and low signal to noise ratio, using the classification accuracy as a test statistic for two sample testing should be performed with extra cautious. Beside these limitations of classification accuracy in inference, and considering the fact that the best predictive model might not be the most informative one~\cite{turnermodel2015}; brain decoding, taken alone, only answers the question of \emph{what} is the most likely label of a given unseen sample~\cite{baehrens2010explain}. This fact is generally known as knowledge extraction gap~\cite{vellido2012making} in the classification context. Therefore, despite the theoretical advantages of MVPA, its practical application to inferences regarding neuroimaging data is limited primarily by a lack of interpretability~\cite{sabuncu2014universal,haynes2015primer,naselaris2015resolving}. Thus far, many efforts have been devoted to filling the knowledge extraction gap of linear and non-linear data modeling methods in different areas such as computer vision~\cite{bach2015pixel}, signal processing~\cite{montavon2013analyzing}, chemometrics~\cite{yu2015classification}, bioinformatics~\cite{hansen2011visual}, and neuroinformatics~\cite{haufe2013interpretation}.

Improving the interpretability of linear brain decoding and associated brain maps is a primary goal in the brain imaging literature~\cite{strother2014stability}. The lack of interpretability of multivariate brain maps is a direct consequence of low signal-to-noise ratios (SNRs), high dimensionality of whole-scalp recordings, high correlations among different dimensions of data, and cross-subject variability~\cite{besserve2007classification,anderson2011common,Brodersen2011Modelbased,lemm2011introduction,langs2011detecting,
varoquaux2012small,kauppi2013decoding,taulu2014novel,varoquaux2014machine,olivetti2014meg,Haufe2014Dimensionality,
haynes2015primer}. At present, two main approaches are proposed to enhance the interpretability of multivariate brain maps: 1) introducing new metrics into the model selection procedure and 2) introducing new penalty terms for regularization to enhance stability selection.

The first approach to improving the interpretability of brain decoding concentrates on the model selection procedure. Model selection is a procedure in which the best values for the hyper-parameters of a model are determined~\cite{lemm2011introduction}. This selection process is generally performed by considering the generalization performance, i.e., the accuracy, of a model as the decisive criterion. \citet{rasmussen2012model} showed that there is a trade-off between the spatial reproducibility and the prediction accuracy of a classifier; therefore, the reliability of maps cannot be assessed merely by focusing on their prediction accuracy. To utilize this finding, they incorporated the spatial reproducibility of brain maps in the model selection procedure. An analogous approach, using a different definition of spatial reproducibility, is proposed by~\citet{conroy2013fast}. Beside spatial reproducibility, the stability of the classifiers~\cite{bousquet2002stability} is another criterion that is used in combination with generalization performance to enhance the interpretability. For example,~\cite{yu2013stability,lim2015estimation} showed that incorporating the stability of models into cross-validation improves the interpretability of the estimated parameters (by linear models).

The second approach to improving the interpretability of brain decoding focuses on the underlying mechanism of regularization. The main idea behind this approach is two-fold: 1) customizing the regularization terms to address the ill-posed nature of brain decoding problems (where the number of samples is much less than the number of features)~\cite{morch1997nonlinear,varoquaux2014machine} and 2) combining the structural and functional prior knowledge with the decoding process so as to enhance stability selection. Group Lasso~\cite{yuan2006model} and total-variation penalty~\cite{tibshirani2005sparsity} are two effective methods using this technique~\cite{xing2014high,rish2014practical}. Sparse penalized discriminant analysis~\cite{grosenick2008interpretable}, group-wise regularization~\cite{van2009interpreting},
randomized Lasso~\cite{varoquaux2012small}, smoothed-sparse logistic regression~\cite{de2012combining}, total-variation L1 penalization~\cite{michel2011total,gramfort2013identifying}, the graph-constrained elastic-net~\cite{grosenick2009whole,grosenick2013interpretable}, and randomized structural sparsity~\cite{wang2015randomized} are examples of brain decoding methods in which regularization techniques are employed to improve stability selection, and thus, the interpretability of brain decoding.

Recently, taking a new approach to the problem, Haufe et al. questioned the interpretability of weights of linear classifiers because of the contribution of noise in the decoding process~\cite{biessmann2012interpretability,haufe2013interpretation,haufe2014parameter}. To address this problem, they proposed a procedure to convert the linear brain decoding models into their equivalent generative models. Their experiments on the simulated and fMRI/EEG data illustrate that, whereas the direct interpretation of classifier weights may cause severe misunderstanding regarding the actual underlying effect, their proposed transformation effectively provides interpretable maps. Despite the theoretical soundness of this method, the major challenge of estimating the empirical covariance matrix of the small sample size neuroimaging data~\cite{engemann2015automated} limits the practical application of this method.

Despite the aforementioned efforts to improve the interpretability of brain decoding, there is still no formal definition for the interpretability of brain decoding in the literature. Therefore, the interpretability of different brain decoding methods are evaluated either qualitatively or indirectly (i.e., by means of an intermediate property). In qualitative evaluation, to show the superiority of one decoding method over the other (or a univariate map), the corresponding brain maps are compared visually in terms of smoothness, sparseness, and coherency using already known facts (see, for example,~\cite{varoquaux2012small,li2015novel}). In the second approach, important factors in interpretability such as spatio-temporal reproducibility are evaluated to indirectly assess the interpretability of results (see, for example,~\cite{langs2011detecting,rasmussen2012model,conroy2013fast,kia2015multitask}). Despite partial effectiveness, there is no general consensus regarding the quantification of these intermediate criteria. For example, in the case of spatial reproducibility, different methods such as correlation~\cite{rasmussen2012model,kia2015multitask}, dice score~\cite{langs2011detecting}, or parameter variability~\cite{haufe2013interpretation,conroy2013fast} are used for quantifying the stability of brain maps, each of which considers different aspects of local or global reproducibility.

With the aim of filling this gap, our contribution in this study is three-fold: 1) Assuming that the true solution of brain decoding is available, we present a theoretical definition of the interpretability. Furthermore, we show that the interpretability can be decomposed into the reproducibility and the representativeness of brain maps. 2) As a proof of the theoretical concepts, we propose a practical heuristic based on event-related fields for quantifying the interpretability of brain maps in MEG decoding scenarios. 3) Finally, we propose the combination of the interpretability and the performance of the brain decoding as a new Pareto optimal multi-objective criterion for model selection. We experimentally show that incorporating the interpretability of the models into the model selection procedure provides more reproducible, more neurophysiologically plausible, and (as a result) more interpretable maps.

\section{Methods}
\label{sec:methods}
\subsection{Notation and Background}
\label{subsec:notations}
Let $\mathcal{X} \in \mathbb{R}^p$ be a manifold in Euclidean space that represents the input space and $\mathcal{Y}\in \mathbb{R}$ be the output space, where $\mathcal{Y}=\Phi^*(\mathcal{X})$. Then, let $S=\{ \textbf{Z} = (\textbf{X},\textbf{Y}) \mid z_1=(x_1,y_1), \dots , z_n=(x_n,y_n)\}$ be a training set of $n$ independently and identically distributed (iid) samples drawn from the joint distribution of $\mathcal{Z} = \mathcal{X} \times \mathcal{Y}$ based on an unknown Borel probability measure $\rho$. In the neuroimaging context, $\textbf{X}$ indicates the trials of brain recording, e.g., fMRI, MEG, or EEG signals, and $\textbf{Y}$ represents the experimental conditions or dependent variables. The goal of brain decoding is to find the function $\Phi_S: \textbf{X} \to \textbf{Y}$ as an estimation of the ideal function $\Phi^*: \mathcal{X} \to \mathcal{Y}$.

In this study, as is a common assumption in the neuroimaging context, we assume the true solution of a brain decoding problem is among the family of linear functions $\mathcal{H}$ ($\Phi^* \in \mathcal{H}$). Therefore, the aim of brain decoding reduces to finding an empirical approximation of $\Phi_S$, indicated by $\hat\Phi$, among all $\Phi \in \mathcal{H}$. This approximation can be obtained by estimating the predictive conditional density $\rho(\textbf{Y} \mid \textbf{X})$ by training a parametric model $\rho(\textbf{Y}\mid \textbf{X}, \Theta)$ (i.e., a likelihood function), where $\Theta$ denotes the parameters of the model. Alternatively, $\Theta$ can be estimated by solving a risk minimization problem:
\begin{eqnarray} \label{eq:max2min}
\begin{split}
\hat{\Theta} = \argmin_{\Theta} \mathcal{L}(\Phi(\textbf{X}),\Phi_S(\textbf{X})+ \lambda \Omega (\Theta)
\end{split}
\end{eqnarray}
where $\mathcal{L}: \textbf{Z} \times \textbf{Z} \to \mathbb{R}^+$ is the loss function, $\Omega:\mathbb{R}^{p}\to \mathbb{R}^+$ is the regularization term, and $\lambda$ is a hyper-parameter that controls the amount of regularization. There are various choices for $\Omega$, each of which reduces the hypothesis space $\mathcal{H}$ to $\mathcal{H}'\subset\mathcal{H}$ by enforcing different prior functional or structural constraints on the parameters of the linear decoding model (see, for example,~\cite{tibshirani1996regression,zou2005regularization,tibshirani2005sparsity,jenatton2009structured}). The amount of regularization $\lambda$ is generally decided using cross-validation or other data perturbation methods in the model selection procedure.

In the neuroimaging context, the estimated parameters of a linear decoding model $\hat{\Theta}$ can be used in the form of a brain map so as to visualize the discriminative neurophysiological effect. Although the magnitude of $\hat{\Theta}$ is affected by the dynamic range of data and the level of regularization, it has no effect on the predictive power and the interpretability of maps. On the other hand, the direction of $\hat{\Theta}$ affects the predictive power and contains information regarding the importance of and relations among predictors. 
This type of relational information is very useful when interpreting brain maps in which the relation between different spatio-temporal independent variables can be used to describe how different brain regions interact over time for a certain cognitive process. Therefore, we refer to the normalized parameter vector of a linear brain decoder in the unit hyper-sphere as a multivariate brain map (MBM); we denote it by $\vec{\Theta}$ where $\vec{\Theta} = \frac{\Theta}{\left \| \Theta \right \|}$ ($\left \| . \right \|$ represents the 2-norm of a vector).

As shown in Eq.~\ref{eq:max2min}, learning occurs using the sampled data. In other words, in the learning paradigm, we attempt to minimize the loss function with respect to $\Phi_S$ (and not $\Phi^*$)~\cite{poggio2002mathematical}. Therefore, all of the implicit assumptions (such as linearity) regarding $\Phi^*$ might not hold on $\Phi_S$, and vice versa (see the supplementary material for a simple illustrative example). The \emph{irreducible error} $\varepsilon$ is the direct consequence of this sampling; it provides a lower bound on the error of a model, where we have:
\begin{eqnarray} \label{eq:estimation}
\Phi_S(\textbf{X}) = \Phi^*(\textbf{X}) + \varepsilon
\end{eqnarray}
The distribution of $\varepsilon$ dictates the type of loss function $\mathcal{L}$ in Eq.~\ref{eq:max2min}. For example, assuming a Gaussian distribution with mean $0$ and variance $\sigma^2$ for $\varepsilon$ implies the least squares loss function~\cite{wu2006complete}.

\subsection{Interpretability of Multivariate Brain Maps: Theoretical Definition}
\label{subsec:interpretability_MBM}
In this section, we introduce a theoretical definition for the interpretability of linear brain decoding models and their associated MBMs. The presented definition remains theoretical, as it is based on a restrictive assumption in practical applications. We assume that the brain decoding problem is linearly separable and that its \emph{unique}, neurophysiologically \emph{plausible}\footnote{Here, neurophysiological plausibility refers to the spatio-temporal chemo-physical constraints of the underlying neural activity that is highly dependent on the acquisition device.} solution, i.e., $\Phi^*$, is available. In this theoretical environment, the goal is to assess the quality of estimated MBMs obtained using different brain decoding methods on a small-sample-size dataset $S$.

Consider a linearly separable brain decoding problem in an ideal scenario where $\varepsilon=0$ and $rank(\textbf{X})=p$. In this case, $\Phi^*$ is linear and its parameters $\Theta^*$ are unique and plausible. The unique parameter vector $\Theta^*$ can be computed as follows:
\begin{eqnarray} \label{eq:ideal_least_square}
\Theta^* = \Sigma_{\textbf{X}}^{-1} \textbf{X}^T \textbf{Y}
\end{eqnarray}

Using $\Theta^*$ as the reference, we define the \emph{strong interpretability} of an MBM as follows:
\begin{mydefs} \label{def:strong_interpretability}
An MBM $\vec\Theta$ associated with a linear function $\Phi$ is ``strongly interpretable" if and only if $\vec\Theta \propto \Theta^*$.
\end{mydefs}

It can be shown that, in practice, the estimated solution of a linear brain problem (using Eq.~\ref{eq:max2min}) is not strongly interpretable because of the inherent limitations of neuroimaging data, such as uncertainty~\cite{aggarwal2009survey} in the input and output space ($\varepsilon \neq 0$), limitations in data acquisition, the high dimensionality of data ($n \ll p$), and the high correlation between predictors ($rank(\textbf{X})<p$). With these limitations in mind, even though linear brain decoders might not be absolutely interpretable, one can argue that some models are more interpretable than others. For example, in the case in which $\Theta^* \propto [0,1]^T$, a linear model where $\hat{\Theta} \propto [0.1,1.2]^T$ can be considered more interpretable than a linear model where $\hat{\Theta} \propto [2,1]^T$. To address this issue, and having in mind the definition of strong-interpretability, our goal is to answer the following question: 

\begin{myprobs} \label{prob:interpretability}
Let $S^1, \dots, S^m$ be $m$ perturbed training sets drawn from $S$ via a certain perturbation scheme such as jackknife, bootstrapping~\cite{efron1979bootstrap}, or cross-validation~\cite{kohavi1995study}. Assume $\vec{\hat\Theta}^1, \dots, \vec{\hat\Theta}^m$ are $m$ MBMs of a certain $\Phi$ (estimated using Eq.~\ref{eq:max2min} for certain $\mathcal{L}$, $\Omega$, and $\lambda$) on the corresponding perturbed training sets. How can we quantify the closeness of $\Phi$ to the strongly-intrepretable solution of brain decoding problem $\Phi^*$?  
\end{myprobs}

To answer this question, considering the uniqueness and the plausibility of $\Phi^*$ as the two main characteristics that convey its strong interpretability, we define the geometrical proximity between $\Phi$ to $\Phi^*$ as a measure for interpretability of $\Phi$.

\begin{mydefs} \label{def:interpretability}
Let $\alpha^j$ ($j = 1,\dots,m$) be the angle between $\vec{\hat\Theta}^j$ and $\vec{\Theta}^*$. The ``interpretability" ($0 \leq \eta_\Phi \leq 1$) of the MBM derived from a linear function $\Phi$ is defined as follows:
\begin{eqnarray} \label{eq:interpretability}
\forall j \in \{1,\dots,m\}, \eta_\Phi = \mathbb{E}_{S}[\cos(\alpha^j)]
\end{eqnarray}
\end{mydefs}

Empirically, the interpretability is the mean of cosine similarities between $\Theta^*$ and MBMs derived from different samplings of the training set. In addition to the fact that employing cosine similarity is a common method for measuring the similarity between vectors, we have another strong motivation for this choice. It can be shown that, for large values of $p$, the distribution of the dot product in the unit hyper-sphere, i.e., the cosine similarity, converges to a normal distribution with $0$ mean and variance of $\frac{1}{p}$, i.e., $\mathcal{N}(0,\sqrt{\frac{1}{p}})$. Due to the small variance for a large enough $p$ values, any similarity significantly larger than zero represents a meaningful similarity between two high dimensional vectors (see the supplementary material for more details about the distribution of cosine similarity). 

In what follows, we demonstrate how the definition of interpretability is geometrically related to the uniqueness and plausibility characteristics of the true solution to brain decoding.

\subsection{Interpretability Decomposition into Reproducibility and Representativeness}
\label{subsec:interpretability_decomposition}
An alternative approach toward quantifying the interpretability of an MBM is to assess its neurophysiological plausibility and uniqueness separately. The high dimensionality and the high correlation between variables are two inherent characteristics of neuroimaging data that negatively affect the uniqueness of the solution of a brain decoding problem. Therefore, a certain configuration of hyper-parameters may result different estimated parameters on different portions of data. Here, we are interested in assessing this variability. Let $\theta^j_i$ be the $i$th ($i = 1,\dots,p$) element of an MBM estimated on the $j$th ($j = 1,\dots,m$) perturbed training set. We define the \emph{main multivariate brain map} as follows:

\begin{mydefs} \label{def:main_map}
The ``main multivariate brain map" $\vec{\Theta}^\mu \in \mathbb{R}^{p}$ of a linear model $\Phi$ is defined as the sum of all estimated MBMs $\vec{\hat{\Theta}}^j$ ($j=1, \dots, m$) on the perturbed training sets $S^j$ in the unit hyper-sphere:
\begin{eqnarray} \label{eq:main_map}
\vec{\Theta}^\mu=\frac{\begin{bmatrix}
\sum_{j=1}^{m}\theta_1^j & \sum_{j=1}^{m}\theta_2^j & \dots & \sum_{j=1}^{m}\theta_p^j
\end{bmatrix}^T}
{\left \| \begin{bmatrix}
\sum_{j=1}^{m}\theta_1^j & \sum_{j=1}^{m}\theta_2^j & \dots & \sum_{j=1}^{m}\theta_p^j
\end{bmatrix}^T \right \|}
\end{eqnarray}
\end{mydefs}

The definition of $\vec{\Theta}^\mu$ is analogous to the main prediction of a learning algorithm~\cite{domingos2000unified}; it provides a reference for quantifying the reproducibility of an MBM as a measure of its uniqueness:

\begin{mydefs} \label{def:reproducibility_MBM}
Let $\vec{\Theta}^\mu$ be the main multivariate brain map of $\Phi$. Then, let $\alpha^j$ be the angle between $\vec{\hat\Theta}^j$ and $\vec{\Theta}^\mu$. The ``reproducibility" $\psi_{\Phi}$ ($0 \leq \psi_{\Phi} \leq 1$) of an MBM derived from a linear function $\Phi$ is defined as follows:
\begin{eqnarray} \label{eq:reproducibility_MBM}
\forall j \in \{1,\dots,m\}, \psi_{\Phi} = \mathbb{E}_{S}[\cos(\alpha^j)]
\end{eqnarray}
\end{mydefs}

In fact, reproducibility provides a measure for quantifying the dispersion of MBMs, computed over different perturbed training sets, from the main multivariate brain map.

In theory, the directional proximity between $\vec\Theta^*$ and the estimated MBM of a linear model provides a measure for plausibility of $\Phi$ that quantifies the coherency between the estimated parameters and the real underlying physiological activities. Here, we define this coherency as the \emph{representativeness} of an MBM.

\begin{mydefs} \label{def:representativeness}
Let $\vec{\Theta}^\mu$ be the main multivariate brain map of $\Phi$. The ``representativeness" ($0 \leq \beta_\Phi \leq 1$) of $\Phi$ is defined as the cosine similarity between $\vec{\Theta}^\mu$ and $\vec{\Theta}^*$:
\begin{eqnarray} \label{eq:representativeness}
\beta_\Phi = \frac{|\vec{\Theta}^\mu.\vec{\Theta}^* |}{\left\| \vec{\Theta}^\mu \right\| \left\| \vec{\Theta}^* \right\|}
\end{eqnarray}
\end{mydefs}

The relationship between the presented definitions for both reproducibility and representativeness and the interpretability can be expressed using the following proposition:

\begin{myprops} \label{theo:interpretability}
$\eta_\Phi = \beta_\Phi \times \psi_\Phi$.
\end{myprops}

See~\ref{sec:interpretability_proof} and Figure~\ref{fig:interpretability_theory} for a proof. Proposition~\ref{theo:interpretability} indicates the interpretability can be decomposed into the representativeness and the reproducibility of a decoding model.

\subsection{A Heuristic for Practical Quantification of Interpretability in Time-Domain MEG decoding}
\label{subsec:interpretability_heuristic}
In practice, it is impossible to evaluate the interpretability, as $\Phi^*$ is unknown. In this study, to provide a practical proof of the mentioned theoretical concepts, we propose the use of contrast event-related fields (cERFs) of MEG data as neurophysiological plausible heuristics for $\Theta^*$ in a binary MEG decoding scenario in the time domain.

The EEG/MEG data are a mixture of several simultaneous stimulus-related and stimulus-unrelated brain activities. In general, unrelated-stimulus brain activities are considered as Gaussian noise with zero mean and variance $\sigma^2$. One popular approach to canceling the noise component is to compute the average of multiple trials. It is expected that the average will converge to the true value of the signal with a variance of $\frac{\sigma^2}{n}$. The result of the averaging process is generally known as ERF in the MEG context; separate interpretation of different ERF components can be performed~\cite{rugg1995electrophysiology}\footnote{The application of the presented heuristic to MEG data can be extended to EEG because of the inherent similarity of the measured neural correlates in these two devices. In the EEG context, the ERF can be replaced by the event-related potential (ERP).}.

Assume $\textbf{X}^+=\{x_i \in \textbf{X} \mid y_i=1 \} \in \mathbb{R}^{n^+ \times p}$ and $\textbf{X}^-=\{x_i \in \textbf{X} \mid y_i=-1 \}\in \mathbb{R}^{n^- \times p}$. Then, the cERF brain map $\vec{\Theta}^{cERF}$ is computed as follows:

\begin{eqnarray} \label{eq:cERF}
\vec{\Theta}^{cERF} = \frac{\frac{1}{n^+} \sum_{x_i \in X^+} x_i - \frac{1}{n^-} \sum_{x_i \in X^-} x_i}{\left \| \frac{1}{n^+} \sum_{x_i \in X^+} x_i - \frac{1}{n^-} \sum_{x_i \in X^-} x_i \right \|}
\end{eqnarray}

Using the core theory presented in~\cite{haufe2013interpretation}, it can be shown that cERF is the equivalent generative model for the least squares solution in a binary time-domain MEG decoding scenario (see~\ref{sec:cERF_generative}). Using $\vec{\Theta}^{cERF}$ as a heuristic for $\vec{\Theta}^*$, the representativeness can be approximated as follows:

\begin{eqnarray} \label{eq:representativeness_estimation1}
\tilde{\beta}_\Phi = \frac{|\vec{\Theta}^\mu.\vec{\Theta}^{cERF} |}{\left\| \vec{\Theta}^\mu \right\| \left\| \vec{\Theta}^{cERF} \right\|}
\end{eqnarray}

Where $\tilde{\beta}_\Phi$ is an approximation of $\beta_\Phi$ and we have:

\begin{eqnarray} \label{eq:representativeness_estimation2}
\beta_\Phi = \Delta_{\beta}\tilde{\beta}_\Phi \pm \sqrt{(1-\tilde{\beta}_\Phi^2)(1-\Delta_{\beta}^2)}
\end{eqnarray}

$\Delta_{\beta}$ represents the cosine similarity between $\vec{\Theta}^*$ and $\vec{\Theta}^{cERF}$ (see Figures~\ref{fig:misrepresentativeness} and~\ref{sec:missrepresentation}). If $\Delta_{\beta} \to 1$ then $\tilde{\beta}_\Phi \to \beta_\Phi$.

In a similar manner, $\vec{\Theta}^{cERF}$ can be used to heuristically approximate the interpretability as follows:

\begin{eqnarray} \label{eq:interpretability_estimation1}
\tilde{\eta}_\Phi = \forall j \in \{1,\dots,m\}, \tilde{\eta}_\Phi = \mathbb{E}_S(\cos(\gamma^j))
\end{eqnarray}

where $\gamma_1, \dots, \gamma_m$ are the angles between $\vec{\hat{\Theta}}^1, \dots, \vec{\hat{\Theta}}^m$ and $\vec{\Theta}^{cERF}$. The following equality represents the relation between $\eta$ and $\tilde{\eta}$ (see Figures~\ref{fig:interpretability_estimation} and~\ref{sec:interpretation_estimation}).

\begin{eqnarray} \label{eq:interpretability_estimation2}
\eta_\Phi = \Delta_{\beta}\tilde{\eta}_\Phi \pm \frac{\sqrt{1-\Delta_{\beta}^2}}{m}(\sin\gamma_1+\dots+\sin\gamma_m)\end{eqnarray}

Again, if $\Delta_{\beta} \to 1$ then $\tilde{\eta}_\Phi \to \eta_\Phi$. Notice that $\Delta_{\beta}$ is independent of the decoding approach used; it only depends on the quality of the heuristic. It can be shown that $\tilde{\eta}_\Phi = \tilde{\beta}_\Phi \times \psi_\Phi$.

Eq.~\ref{eq:interpretability_estimation2} shows that the choice of heuristic has a direct effect on the approximation of interpretability and that an inappropriate selection of the heuristic yields a very poor estimation of interpretability because of the destructive contribution of $\Delta_{\beta}$. Therefore, the choice of heuristic should be carefully justified based on accepted and well-defined facts regarding the nature of the collected data (see the supplementary material for the experimental investigation of the limitations of the proposed heuristic).

\subsection{Incorporating the Interpretability into Model Selection}
\label{subsec:interpretability_model_selection}
The procedure for evaluating the performance of a model so as to choose the best values for hyper-parameters is known as \emph{model selection}~\cite{hastie2009elements}. This procedure generally involves numerical optimization of the model selection criterion. The most common model selection criterion is based on an estimator of generalization performance, i.e., the predictive power. In the context of brain decoding, especially when the interpretability of brain maps matters, employing only the predictive power of the decoding model in model selection is problematic in terms of interpretability~\cite{gramfort2012beyond,rasmussen2012model,conroy2013fast}. Here, we propose a multi-objective criterion for model selection that takes into account both prediction accuracy and MBM interpretability.

Let $\tilde{\eta}_\Phi$ and $\delta_\Phi$ be the approximated interpretability and the generalization performance of a linear function $\Phi$, respectively. We propose the use of the \emph{scalarization} technique~\cite{caramia2008multi} for combining $\tilde{\eta}_\Phi$ and $\delta_\Phi$ into one scalar $0 \leq \zeta(\Phi) \leq 1$ as follows:
\begin{eqnarray} \label{eq:plausibility}
\zeta_\Phi =
\left\{\begin{matrix}
\frac{\omega_1 \tilde{\eta}_\Phi + \omega_2 \delta_\Phi}{\omega_1 + \omega_2}  & \delta_\Phi \geq \kappa \\
0 & \delta_\Phi < \kappa
\end{matrix}\right.
\end{eqnarray}

where $\omega_1$ and $\omega_2$ are weights that specify the level of importance of the interpretability and the performance of the model, respectively. $\kappa$ is a threshold on the performance that filters out solutions with poor performance. In classification scenarios, $\kappa$ can be set by adding a small safe interval to the chance level of classification.

It can be shown that the hyper-parameters of a model $\Phi$ are optimized based on $\zeta_\Phi$ are Pareto optimal~\cite{marler2004survey}. In other words, there exist no other $\Phi'$ for which we obtain both $\tilde{\eta}_{\Phi'}>\tilde{\eta}_\Phi$ and $\delta_{\Phi'}>\delta_\Phi$. We expect that optimizing the hyper-parameters of the model based on $\zeta_\Phi$, rather only $\delta_\Phi$, yields more informative MBMs.

\subsection{Experimental Materials}
\label{subsec:materials}
\subsubsection{Toy Dataset}
\label{subsubsec:toy_data}
To illustrate the importance of integrating the interpretability of brain decoding with the model selection procedure, we use simple 2-dimensional toy data presented in~\cite{haufe2013interpretation}. Assume that the true underlying generative function $\Phi^*$ is defined by

\begin{align*}
\mathcal{Y}=\Phi^*(\mathcal{X})=\left\{\begin{matrix}
1 & \quad if \quad \textit{x}_1 = 1.5 \\
-1 & \quad if \quad \textit{x}_1 = -1.5
\end{matrix}\right.
\end{align*}

where $\mathcal{X} \in \{ [1.5,0]^T, [-1.5,0]^T\}$; and $x_1$ and $x_2$ represent the first and the second dimension of the data, respectively. Furthermore, assume the data is contaminated by Gaussian noise with co-variance $\Sigma=\begin{bmatrix} 1.02 & -0.3\\ -0.3 & 0.15 \end{bmatrix}$. Figure~\ref{fig:simulation_study_1_data} shows the distribution of the noisy data.

\subsubsection{MEG Data}
\label{subsubsec:meg_data}
In this study, we use the MEG dataset presented in~\cite{10.3389/fnhum.2011.00076}\footnote{The full dataset is publicly available at \url{ftp://ftp.mrc-cbu.cam.ac.uk/personal/rik.henson/wakemandg_hensonrn/}}. This dataset was also used for the DecMeg2014 competition\footnote{The competition data are available at \url{http://www.kaggle.com/c/decoding-the-human-brain}}. In this dataset, visual stimuli consisting of famous faces, unfamiliar faces, and scrambled faces are presented to $16$ subjects and fMRI, EEG, and MEG signals are recorded. In this study, we are only interested in MEG recordings. The MEG data were recorded using a VectorView system (Elekta Neuromag, Helsinki, Finland) with a magnetometer and two orthogonal planar gradiometers located at 102 positions in
a hemispherical array in a light Elekta-Neuromag magnetically shielded room.

Three major reasons motivated the choice of this dataset: 1) It is publicly available. 2) The spatio-temporal dynamic of the MEG signal for face vs. scramble stimuli has been well studied. The event-related potential analysis of EEG/MEG shows that $N170$ occurs $130-200 ms$ after stimulus presentation and reflects the neural processing of faces~\cite{bentin1996electrophysiological,10.3389/fnhum.2011.00076}. Therefore, the $N170$ component can be considered the ground truth for our analysis. 3) In the literature, non-parametric mass-univariate analysis such as cluster-based permutation tests is unable to identify narrowly distributed effects in space and time (e.g., an $N170$ component)~\cite{groppe2011amass,groppe2011bmass}. These facts motivate us to employ multivariate approaches that are more sensitive to these effects.

As in \cite{olivetti2014meg}, we created a balanced face vs. scrambled MEG dataset by randomly drawing from the trials of unscrambled (famous or unfamiliar) faces and scrambled faces in equal number. The samples in the face and scrambled face categories are labeled as $1$ and $-1$, respectively. The raw data is high-pass filtered at $1 Hz$, down-sampled to $250 Hz$, and trimmed from $200 ms$ before the stimulus onset to $800 ms$ after the stimulus. Thus, each trial has $250$ time-points for each of the $306$ MEG sensors ($102$ magnetometers and $204$ planar gradiometers)\footnote{The preprocessing scripts in python and MATLAB are available at: \url{https://github.com/FBK-NILab/DecMeg2014/}}. To create the feature vector of each sample, we pooled all of the temporal data of $306$ MEG sensors into one vector (i.e., we have $p = 250 \times 306 = 76500$ features for each sample). Before training the classifier, all of the features are standardized to have a mean of $0$ and standard-deviation of $1$.

\subsection{Classification and Evaluation}
\label{subsubsec:classification_evaluation}
In all experiments, a least squares classifier with L1-penalization, i.e., Lasso~\cite{tibshirani1996regression}, is used for decoding. Lasso is a very popular classification method in the context of brain decoding, mainly because of its sparsity assumption. The choice of Lasso helps us to better illustrate the importance of including the interpretability in the model selection. Lasso solves the following optimization problem:
\begin{eqnarray} \label{eq:lasso}
\hat{\Theta} = \argmin_{\Theta} \left\| \Phi(\textbf{X})-\Phi_S(\textbf{X}) \right\|_2^2 + \lambda \left \| \Theta \right\|_1
\end{eqnarray}

where $\lambda$ is the hyper-parameter that specifies the level of regularization. Therefore, the aim of the model selection is to find the best value for $\lambda$. In this study, we try to find the best regularization parameter value among $\lambda=\{0.001, 0.01, 0.1, 1, 10, 50, 100, 250, 500, 1000, 5000, 10000, 15000, 25000, 50000\}$.

We use the out-of-bag (OOB)~\cite{tibshirani1996bias, wolpert1999efficient, breiman2001random} method for computing $\delta_\Phi$, $\psi_\Phi$, $\tilde{\beta}_\Phi$, $\tilde{\eta}_\Phi$, and $\zeta_\Phi$ for different values of $\lambda$. In OOB, given a training set $(\textbf{X},\textbf{Y})$, $m$ replications of bootstrap~\cite{efron1979bootstrap} are used to create perturbed training sets (we set $m = 50$)~\footnote{The MATLAB code used for experiments is available at \url{https://github.com/smkia/interpretability/}}. In all of our experiments, we set $\omega_1 = \omega_2 = 1$ and $\kappa=0.6$ in the computation of $\zeta_\Phi$. Furthermore, we set $\delta_\Phi=1-EPE$ where EPE indicates the expected prediction error; it is computed using the procedure explained in~\ref{sec:bias_variance_computation}. Employing OOB provides the possibility of computing the bias and variance of the model as contributing factors in EPE.

To investigate the behavior of the proposed model selection criterion, we benchmark it against the commonly used performance criterion in the single-subject decoding scenario. Assuming $(\textbf{X}_i,\textbf{Y}_i)$ for $i=1,\dots,16$ are MEG trial/label pairs for subject $i$, we separately train a Lasso model for each subject to estimate the parameter of the linear function $\hat{\Phi}_i$, where $\textbf{Y}_i = \textbf{X}_i\hat{\Theta}_i$. Let $\hat{\Phi}_i^{\delta}$ and $\hat{\Phi}_i^{\zeta}$ represent the optimized solution based on $\delta_\Phi$ and $\zeta_\Phi$, respectively. We denote the MBM associated with $\hat{\Phi}_i^{\delta}$ and $\hat{\Phi}_i^{\zeta}$ by $\vec{\hat{\Theta}}_i^{\delta}$ and $\vec{\hat{\Theta}}_i^{\zeta}$, respectively. Therefore, for each subject, we compare the resulting decoders and MBMs computed based on these two model selection criteria.
\section{Results}
\label{sec:results}
\subsection{Performance-Interpretability Dilemma: A Toy Example}
\label{subsec:simulation_study}
In the definition of $\Phi^*$ on the toy dataset discussed in Section~\ref{subsubsec:toy_data}, $x_1$ is the decisive variable and $x_2$ has no effect on the classification of the data into target classes. Therefore, excluding the effect of noise and based on the theory of the maximal margin classifier~\cite{vapnik1982estimation,vapnik2013nature}, $\vec{\Theta}^* \propto [1,0]^T$ is the true solution to the decoding problem. By accounting for the effect of noise and solving the decoding problem in $(\textbf{X},\textbf{Y})$ space, we have $\vec{\Theta} \propto [\frac{1}{\sqrt(5)},\frac{2}{\sqrt(5)}]^T$ as the parameter of the linear classifier. Although the estimated parameters on the noisy data provide the best generalization performance for the noisy samples, any attempt to interpret this solution fails, as it yields the wrong conclusion with respect to the ground truth (it says $x_2$ has twice the influence of $x_1$ on the results, whereas it has no effect). This simple experiment shows that the most accurate model is not always the most interpretable model, primarily because the contribution of the noise in the decoding process~\cite{haufe2013interpretation}. On the other hand, the true solution of the problem $\vec{\Theta}^*$ does not provide the best generalization performance for the noisy data.

\begin{figure*}
         \centering
         \begin{subfigure}[h]{0.75\textwidth}
         	\includegraphics[width=\textwidth]{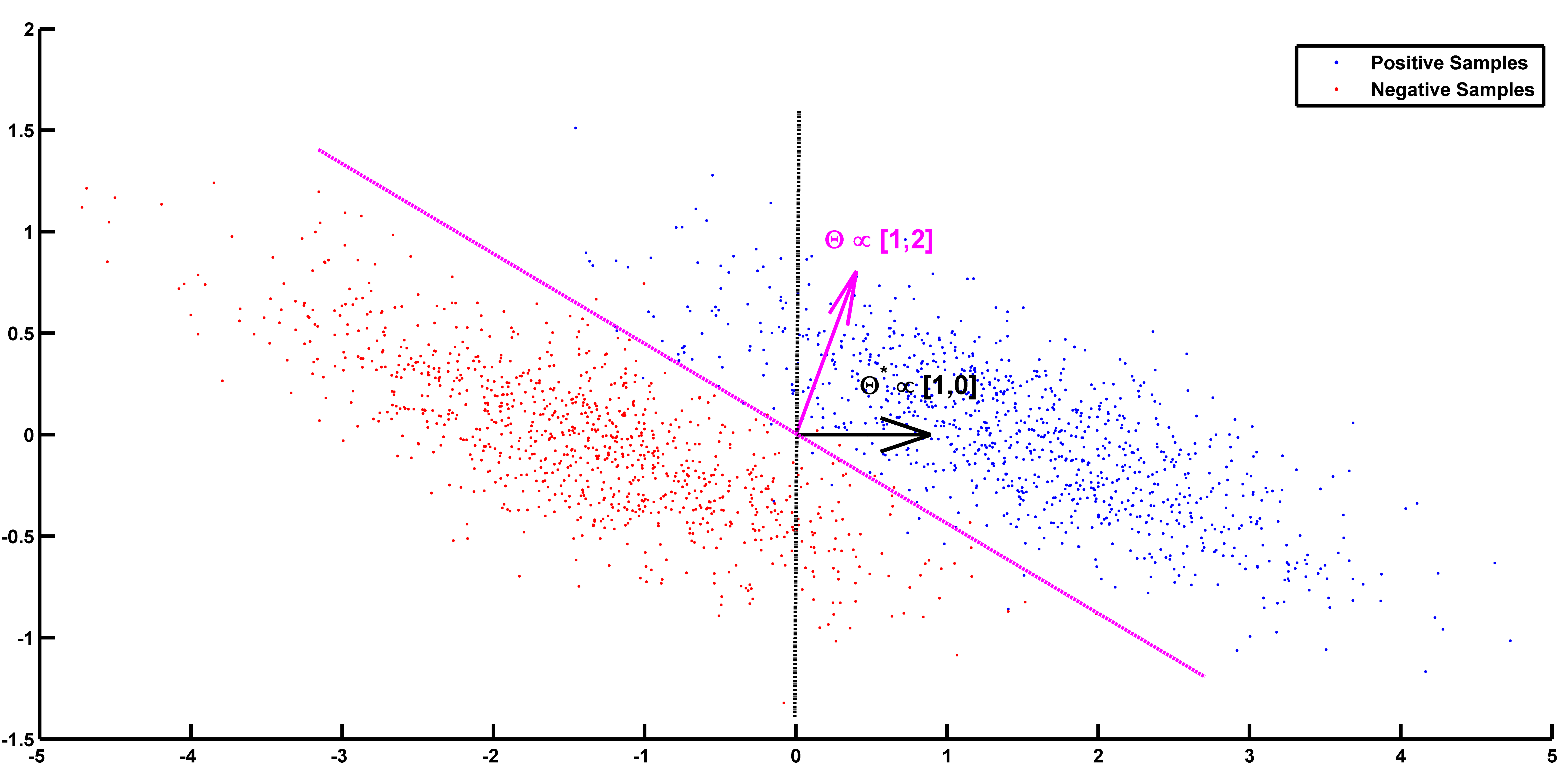}
         \end{subfigure}
         \caption{Noisy samples of toy data. The black line shows the true separator based on the generative model ($\Phi^*$). The magenta line shows the most accurate classification solution. Because of the contribution of noise, any interpretation of the parameters of the most accurate classifier yields a misleading conclusion with respect to the true underlying phenomenon~\cite{haufe2013interpretation}.}
         \label{fig:simulation_study_1_data}
\end{figure*}

To illustrate the effect of incorporating the interpretability in the model selection, a Lasso model with different $\lambda$ values is used for classifying the toy data. In this case, because $\vec{\Theta}^*$ is known, the exact value of interpretability can be computed using Eq.~\ref{eq:interpretability}. Table~\ref{tab:simulation_study_1_results} compares the resultant performance and interpretability from Lasso. Lasso achieves its highest performance ($\delta_\Phi = 0.9884$) at $\lambda=10$ with $\vec{\hat{\Theta}} \propto [0.4636, 0.8660]^T$ (indicated by the magenta line in Figure~\ref{fig:simulation_study_1_data}). Despite having the highest performance, this solution suffers from a lack of interpretability ($\eta_\Phi=0.4484$). By increasing $\lambda$, the interpretability of the model increases. For $\lambda = 500, 1000$ the model reaches its highest interpretability by compensating for $0.06$ of its performance. This observation highlights two main points:

\begin{enumerate}
\item In the case of noisy data, the interpretability of a decoding model is incoherent with its performance. Thus, optimizing the parameter of the model based on its performance does not necessarily improve its interpretability. This observation confirms the previous finding by~\citet{rasmussen2012model} regarding the trade-off between the spatial reproducibility (as a measure for the interpretability of a model) and the prediction accuracy in brain decoding.
\item If the right criterion is used in the model selection, employing proper regularization technique (sparsity prior, in this case) provides more interpretability for the decoding models.
\end{enumerate}

\begin{table}[]
\centering
\caption{Comparison between $\delta_\Phi$, $\eta_\Phi$, and $\zeta_\Phi$ for different $\lambda$ values on the toy 2D example shows the performance-interpretability dilemma, in which the most accurate classifier is not the most interpretable one.}
\label{tab:simulation_study_1_results}
\tiny\setlength{\tabcolsep}{2.5pt}
\begin{tabular}{@{}l|ccccccccccc@{}}
\toprule
$\lambda$                    & 0                                                & 0.001                                             & 0.01                                             & 0.1                                              & 1                                                & 10                                               & 50                                               & 100                                             & 250                                   & 500                                   & 1000                                  \\ \midrule
$\delta(\Phi)$                          & 0.9883                                           & 0.9883                                            & 0.9883                                           & 0.9883                                           & 0.9883                                           & \textbf{0.9884}                                  & 0.9880                                           & 0.9840                                          & 0.9310                                & 0.9292                                & 0.9292                                \\
$\eta(\Phi)$           & 0.4391                                           & 0.4391                                            & 0.4391                                           & 0.4392                                           & 0.4400                                           & 0.4484                                           & 0.4921                                           & 0.5845                                          & 0.9968                                & \textbf{1}                            & \textbf{1}                            \\
$\zeta(\Phi)$          & 0.7137                                           & 0.7137                                            & 0.7137                                           & 0.7137                                           & 0.7142                                           & 0.7184                                           & 0.7400                                           & 0.7842                                          & 0.9639                                & \textbf{0.9646}                       & \textbf{0.9646}                       \\
$\vec{\hat{\Theta}} \propto$ & $\begin{bmatrix} 0.4520 \\ 0.8920 \end{bmatrix}$ & $\begin{bmatrix} 0.4520 \\ 0.8920  \end{bmatrix}$ & $\begin{bmatrix} 0.4520 \\ 0.8920 \end{bmatrix}$ & $\begin{bmatrix} 0.4521 \\ 0.8919 \end{bmatrix}$ & $\begin{bmatrix} 0.4532 \\ 0.8914 \end{bmatrix}$ & $\begin{bmatrix} 0.4636 \\ 0.8660 \end{bmatrix}$ & $\begin{bmatrix} 0.4883 \\ 0.8727 \end{bmatrix}$ & $\begin{bmatrix}0.5800 \\ 0.8146 \end{bmatrix}$ & $\begin{bmatrix}0.99 \\ 0.02 \end{bmatrix}$ & $\begin{bmatrix}1 \\ 0 \end{bmatrix}$ & $\begin{bmatrix}1 \\ 0 \end{bmatrix}$ \\ \bottomrule
\end{tabular}
\end{table}

\subsection{Mass-Univariate Hypothesis Testing on MEG Data}
\label{subsec:mass_univariate_MEG}
Results show that non-parametric mass-univariate analysis is unable to detect narrowly distributed effects in space and time (e.g., an $N170$ component)~\cite{groppe2011amass,groppe2011bmass}. To illustrate the advantage of the proposed decoding framework for spotting these effects, we performed a non-parametric cluster-based permutation test~\cite{maris2007nonparametric} on our MEG dataset using Fieldtrip toolbox~\cite{oostenveld2010fieldtrip}. In a single subject analysis scenario, we considered the trials of MEG recordings as the unit of observation in a between-trials experiment. Independent-samples t-statistics are used as the statistics for evaluating the effect at the sample level and to construct spatio-temporal clusters. The maximum of the cluster-level summed t-value is used for the cluster level statistics; the significance probability is computed using a Monte Carlo method. The minimum number of neighboring channels for computing the clusters is set to $2$. Considering $0.025$ as the two-sided threshold for testing the significance level and repeating the procedure separately for magnetometers and combined-gradiometers, no significant result is found for any of the $16$ subjects. This result motivates the search for more sensitive (and, at the same time, more interpretable) alternatives for hypothesis testing.

\subsection{Single-Subject Decoding on MEG Data}
\label{subsec:real_data_sensor_MEG}
\begin{figure}
         \centering
         \begin{subfigure}[t]{1\textwidth}
                 \includegraphics[width=\textwidth]{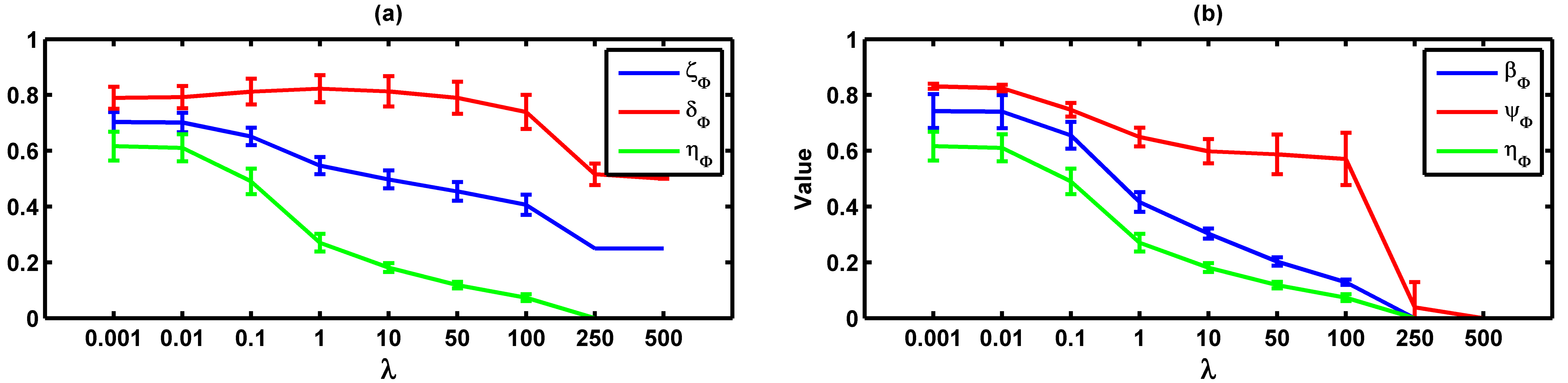}
         \end{subfigure}
         \caption{(a) Mean and standard-deviation of the performance, interpretability, and plausibility of Lasso over 16 subjects. The performance and interpretability become incoherent as $\lambda$ increases. (b) Mean and standard-deviation of the reproducibility, representativeness, and interpretability of Lasso over $16$ subjects. The interpretability declines because of the decrease in both reproducibility and representativeness.}
         \label{fig:MEG_sensor_errorbar_plots}
\end{figure}

In this experiment, we aim to compare the multivariate brain maps of brain decoding models when $\delta_\Phi$ and $\zeta_\Phi$ are used as the criteria for model selection. Figure~\ref{fig:MEG_sensor_errorbar_plots}(a) represents the mean and standard-deviation of the performance and interpretability of Lasso across $16$ subjects for different $\lambda$ values. The performance and interpretability curves further illustrate the performance-interpretability dilemma in the single-subject decoding scenario in which increasing the performance delivers less interpretability. The average performance across subjects is improved when $\lambda$ approaches $1$, but on the other side, the reproducibility and the representativeness of models declines significantly [see Figure~\ref{fig:MEG_sensor_errorbar_plots}(b)].

\begin{figure}
         \centering
         \begin{subfigure}[t]{0.75\textwidth}
                 \includegraphics[width=\textwidth]{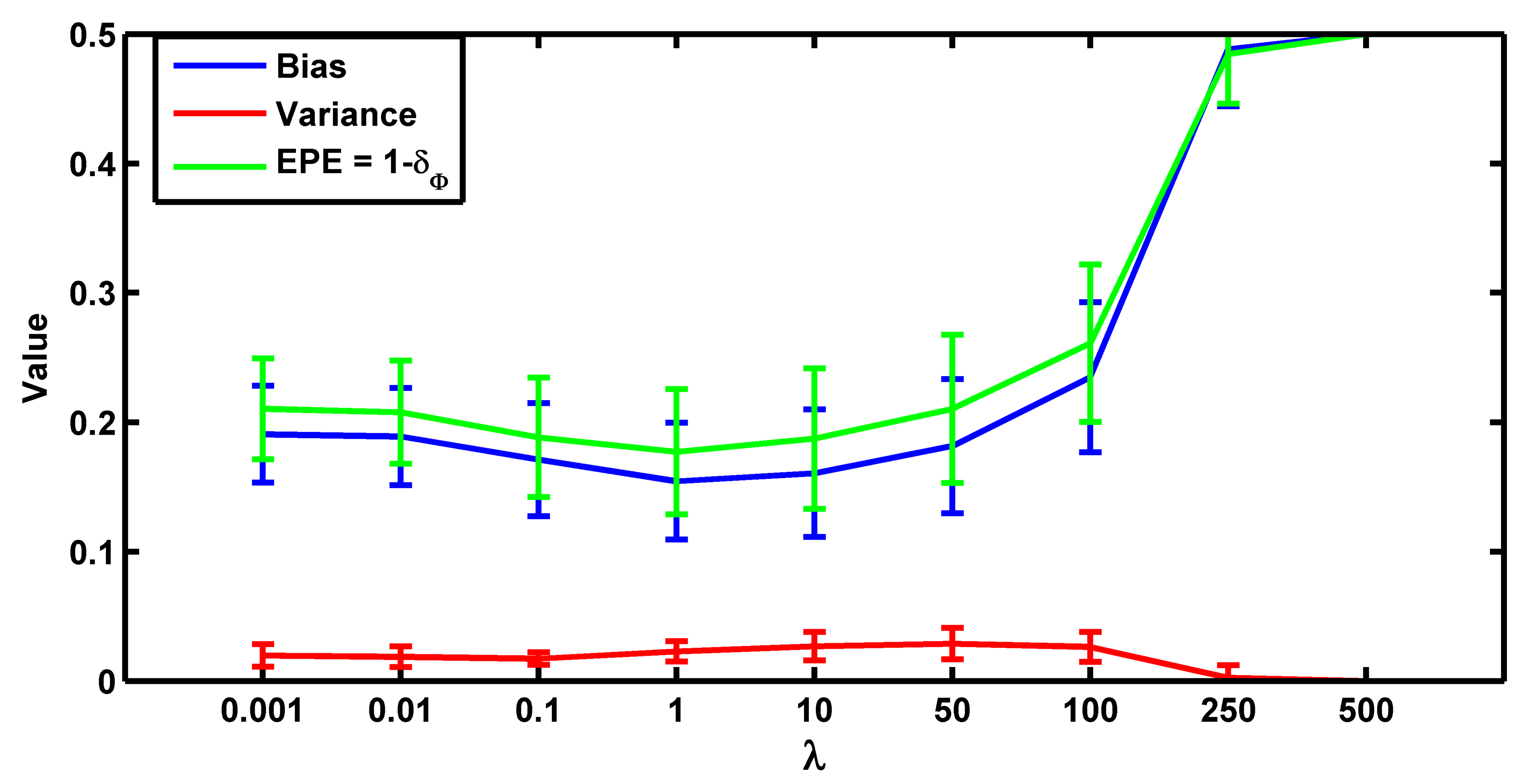}
         \end{subfigure}
         \caption{Mean and standard-deviation of the bias, variance, and EPE of Lasso over 16 subjects. The effect of variance on the EPE is overwhelmed by bias.}
         \label{fig:MEG_sensor_bias_variance}
\end{figure}

One possible reason behind the performance-interpretability dilemma is illustrated in Figure~\ref{fig:MEG_sensor_bias_variance}. The figure shows the mean and standard deviation of bias, variance, and EPE of Lasso across $16$ subjects. The plot proposes that the effect of variance is overwhelmed by bias in the computation of EPE, where the best performance (minimum EPE) at $\lambda = 1$ has the lowest bias, its variance is higher than for $\lambda = 0.001, 0.01, 0.1$. While this tiny increase in the variance is not reflected in EPE but Figure~\ref{fig:MEG_sensor_errorbar_plots}(b) shows a significant effect on the reproducibility of the model.

Table~\ref{tab:MEG_Sensor_Results} summarizes the performance, reproducibility, representativeness, and interpretability of $\hat{\Phi}_i^{\delta}$ and $\hat{\Phi}_i^{\zeta}$ for $16$ subjects. The average result over 16 subjects shows that employing $\zeta_{\Phi}$ instead of $\delta_{\Phi}$ in model selection provides significantly higher reproducibility, representativeness, and (as a result) interpretability compensating for $0.04$ of performance.

\begin{table}[]
\centering
\caption{The performance, reproducibility, representativeness, and interpretability of $\hat{\Phi}_i^{\delta}$ and $\hat{\Phi}_i^{\zeta}$ over 16 subjects.}
\label{tab:MEG_Sensor_Results}
\tiny\setlength{\tabcolsep}{2pt}
\begin{tabular}{|c|ccccc|ccccc|}
\hline
\multirow{2}{*}{Subj} & \multicolumn{5}{c|}{Criterion: $\delta_\Phi$} & \multicolumn{5}{c|}{Criterion: $\zeta_\Phi$} \\ \cline{2-11}
 & $\delta_\Phi$ & $\zeta_\Phi$ & $\tilde{\eta}_\Phi$ & $\tilde{\beta}_\Phi$ & $\psi_\Phi$ & $\delta_\Phi$ & $\zeta_\Phi$ & $\tilde{\eta}_\Phi$ & $\tilde{\beta}_\Phi$ & $\psi_\Phi$ \\ \hline
1 & 0.81 & 0.53 & 0.26 & 0.42 & 0.62 & 0.78 & 0.70 & 0.63 & 0.76 & 0.83 \\
2 & 0.80 & 0.70 & 0.60 & 0.72 & 0.83 & 0.80 & 0.70 & 0.60 & 0.72 & 0.83 \\
3 & 0.81 & 0.63 & 0.45 & 0.64 & 0.71 & 0.78 & 0.71 & 0.64 & 0.78 & 0.83 \\
4 & 0.84 & 0.52 & 0.20 & 0.31 & 0.66 & 0.76 & 0.70 & 0.64 & 0.77 & 0.83 \\
5 & 0.80 & 0.54 & 0.29 & 0.44 & 0.65 & 0.78 & 0.69 & 0.61 & 0.73 & 0.83 \\
6 & 0.79 & 0.52 & 0.24 & 0.39 & 0.63 & 0.74 & 0.67 & 0.61 & 0.74 & 0.82 \\
7 & 0.84 & 0.55 & 0.27 & 0.40 & 0.66 & 0.81 & 0.70 & 0.59 & 0.71 & 0.84 \\
8 & 0.87 & 0.55 & 0.24 & 0.35 & 0.68 & 0.85 & 0.68 & 0.52 & 0.61 & 0.84 \\
9 & 0.80 & 0.55 & 0.31 & 0.46 & 0.67 & 0.77 & 0.67 & 0.57 & 0.69 & 0.82 \\
10 & 0.79 & 0.53 & 0.26 & 0.41 & 0.64 & 0.77 & 0.68 & 0.58 & 0.70 & 0.83 \\
11 & 0.74 & 0.65 & 0.56 & 0.68 & 0.82 & 0.74 & 0.65 & 0.56 & 0.68 & 0.82 \\
12 & 0.80 & 0.55 & 0.29 & 0.46 & 0.64 & 0.79 & 0.70 & 0.61 & 0.74 & 0.83 \\
13 & 0.83 & 0.50 & 0.18 & 0.29 & 0.61 & 0.77 & 0.70 & 0.63 & 0.76 & 0.82 \\
14 & 0.90 & 0.58 & 0.27 & 0.39 & 0.68 & 0.81 & 0.78 & 0.74 & 0.89 & 0.84 \\
15 & 0.92 & 0.63 & 0.34 & 0.48 & 0.71 & 0.89 & 0.78 & 0.66 & 0.77 & 0.86 \\
16 & 0.87 & 0.55 & 0.23 & 0.37 & 0.62 & 0.81 & 0.74 & 0.67 & 0.81 & 0.83 \\ \hline
Mean & \textbf{0.83}$\pm$\textbf{0.05} & $0.57\pm0.05$ & $0.31\pm0.12$ & $0.45\pm0.13$ & $0.68\pm0.07$ & $0.79\pm0.04$ & \textbf{0.70}$\pm$\textbf{0.04} & \textbf{0.62}$\pm$\textbf{0.05} & \textbf{0.74}$\pm$\textbf{0.06} & \textbf{0.83}$\pm$\textbf{0.01} \\ \hline
\end{tabular}
\end{table}

\begin{figure}
         \centering
         \begin{subfigure}[h]{1\textwidth}
                 \includegraphics[width=\textwidth]{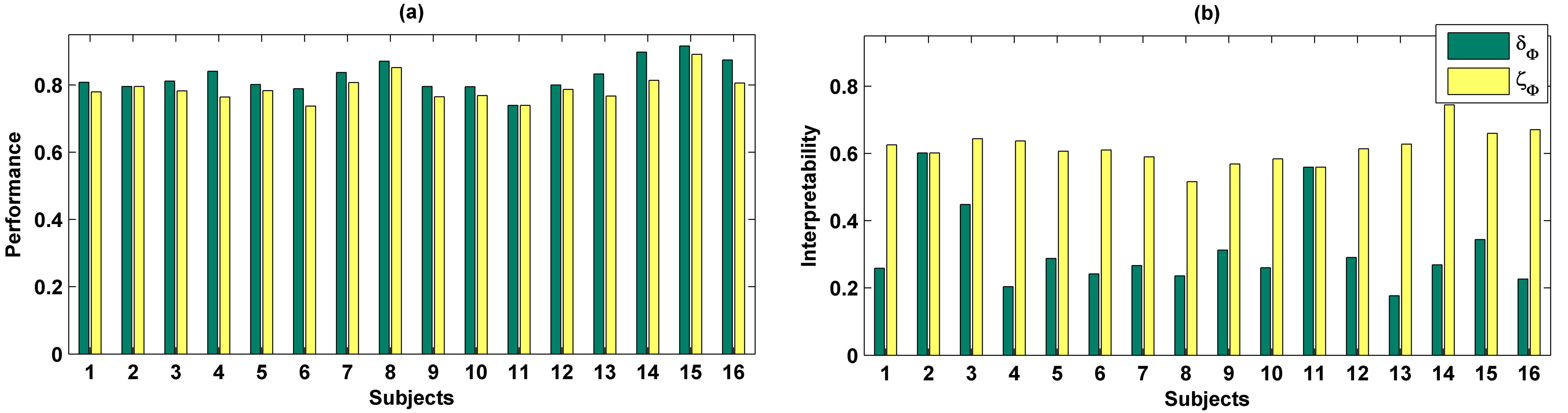}
         \end{subfigure}
         \caption{a) Comparison between performance of $\hat{\Phi}_i^{\delta}$ and $\hat{\Phi}_i^{\zeta}$. Adopting $\zeta_\Phi$ instead of $\delta_\Phi$ in model selection yields (on average) $0.04$ less accurate classifiers over 16 subjects. b) Comparison between interpretability of $\hat{\Phi}_i^{\delta}$ and $\hat{\Phi}_i^{\zeta}$. Adopting $\zeta_\Phi$ instead of $\delta_\Phi$ in model selection yields on average $0.31$ more interpretable classifiers over 16 subjects.}
         \label{fig:MEG_sensor_performanceVSinterpretability}
\end{figure}

These results are further analyzed in Figure~\ref{fig:MEG_sensor_performanceVSinterpretability} where $\hat{\Phi}_i^{\delta}$ and $\hat{\Phi}_i^{\zeta}$ are compared subject-wise in terms of their performance and interpretability. The comparison shows that adopting $\zeta_\Phi$ instead of $\delta_\Phi$ as the criterion for model selection yields significantly better interpretable models by compensating a negligible degree of performance in $14$ out of $16$ subjects. Figure~\ref{fig:MEG_sensor_performanceVSinterpretability}(a) shows that employing $\delta_\Phi$ provides on average slightly higher accurate models (Wilcoxon rank sum test p-value$ = 0.012$) across subjects ($0.83\pm0.05$) than using $\zeta_\Phi$ ($0.79\pm0.04$). On the other side, Figure~\ref{fig:MEG_sensor_performanceVSinterpretability}(b) shows that employing $\zeta_\Phi$ and compensating by $0.04$ in the performance provides (on average) substantially higher (Wilcoxon rank sum test p-value$= 5.6 \times 10^{-6}$) interpretability across subjects ($0.62 \pm 0.05$) compared to $\delta_\Phi$ ($0.31 \pm 0.12$). For example, in the case of subject 1 (see table~\ref{tab:MEG_Sensor_Results}), using $\delta_\Phi$ in model selection to select the best $\lambda$ value for the Lasso model yields a model with $\delta_\Phi=0.81$ and $\tilde{\eta}_\Phi=0.26$. In contrast, using $\zeta_\Phi$ provides a model with $\delta_\Phi=0.78$ and $\tilde{\eta}_\Phi=0.63$.

\begin{figure}
         \centering
         \begin{subfigure}[t]{1\textwidth}
                 \includegraphics[width=\textwidth]{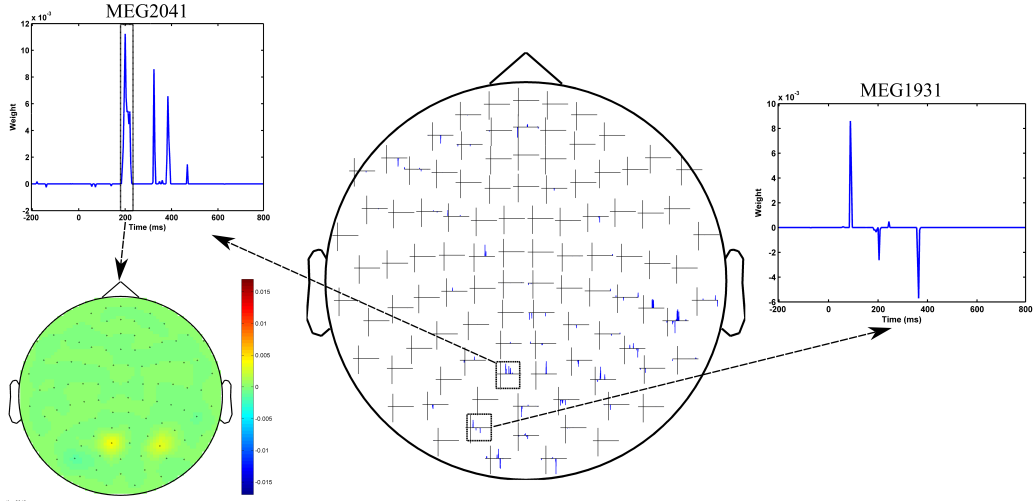}
                 \caption{Spatio-temporal pattern of $\vec{\hat{\Theta}}_1^{\delta}$.}
         		 \label{subfig:subj1_PM_Lasso}
         \end{subfigure}
         \centering
         \begin{subfigure}[t]{1\textwidth}
                 \includegraphics[width=\textwidth]{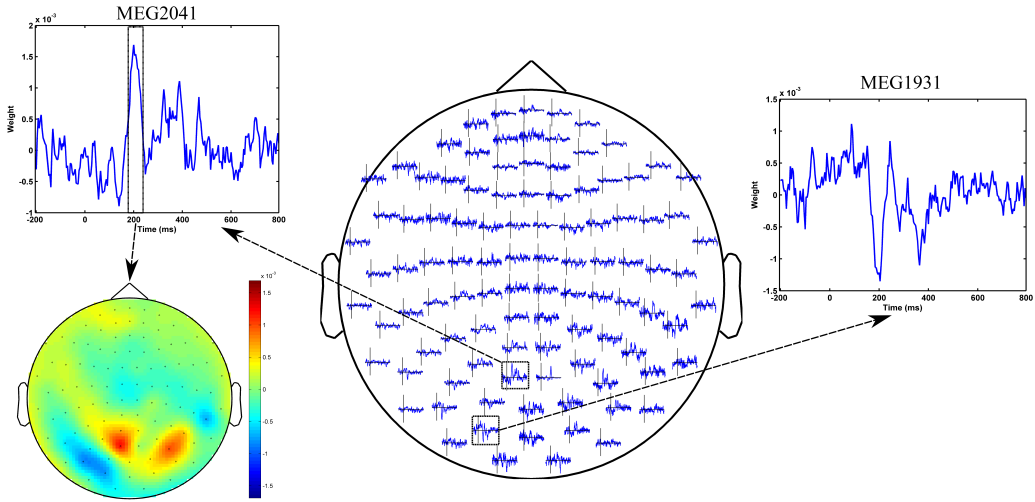}
                 \caption{Spatio-temporal pattern of $\vec{\hat{\Theta}}_1^{\zeta}$.}
         		 \label{subfig:subj1_IM_Lasso}
         \end{subfigure}
         \caption{Comparison between spatio-temporal multivariate maps of the most accurate (~\ref{subfig:subj1_PM_Lasso}) and the most interpretable (~\ref{subfig:subj1_IM_Lasso}) classifiers for Subject 1. $\vec{\hat{\Theta}}_1^{\zeta}$ provides more spatio-temporal representativeness of the N170 effect than $\vec{\hat{\Theta}}_1^{\delta}$.}
         \label{fig:subj1_PMvsIM}
\end{figure}

\begin{figure}
         \centering
         \begin{subfigure}[h]{1\textwidth}
                 \includegraphics[width=\textwidth]{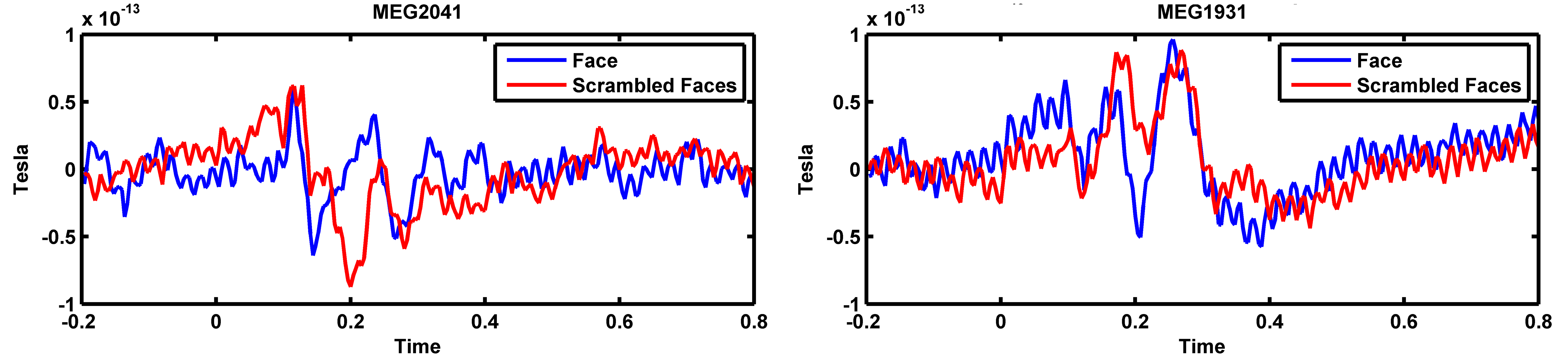}
         \end{subfigure}
         \caption{Event related fields (ERFs) of face and scrambled face samples for MEG2041 and MEG1931 sensors.}
         \label{fig:MEG_sensor_FacevsScrambled_ERP}
\end{figure}

The advantage of the exchange between the performance and the interpretability can be seen in the quality of MBMs. Figure~\ref{subfig:subj1_PM_Lasso} and~\ref{subfig:subj1_IM_Lasso} show $\vec{\hat{\Theta}}_1^{\delta}$ and $\vec{\hat{\Theta}}_1^{\zeta}$ of subject 1, i.e., the spatio-temporal multivariate maps of the Lasso models with maximum values of $\delta_\Phi$ and $\zeta_\Phi$, respectively. The maps are plotted for 102 magnetometer sensors. In each case, the time course of weights of classifiers associated with the MEG2041 and MEG1931 sensors are plotted. Furthermore, the topographic maps represent the spatial patterns of weights averaged between $184 ms$ and $236 ms$ after stimulus onset\footnote{The bounds of colorbars are symmetrized based on the maximum absolute value of parameters}. While $\vec{\hat{\Theta}}_1^{\delta}$ is sparse in time and space, it fails to accurately represent the spatio-temporal dynamic of the N170 component. Furthermore, the multicollinearity problem arising from the correlation between the time course of the MEG2041 and MEG1931 sensors causes extra attenuation of the N170 effect in the MEG1931 sensor. Therefore, the model is unable to capture the spatial pattern of the dipole in the posterior area. In contrast, $\vec{\hat{\Theta}}_1^{\zeta}$ represents the dynamic of the N170 component in time (see Figure~\ref{fig:MEG_sensor_FacevsScrambled_ERP}). In addition, it also shows the spatial pattern of two dipoles in the posterior and temporal areas. In summary, $\vec{\hat{\Theta}}_1^{\zeta}$ suggests a more representative pattern of the underlying neurophysiological effect than $\vec{\hat{\Theta}}_1^{\delta}$.

\begin{figure}
         \centering
         \begin{subfigure}[t]{1\textwidth}
                 \includegraphics[width=\textwidth]{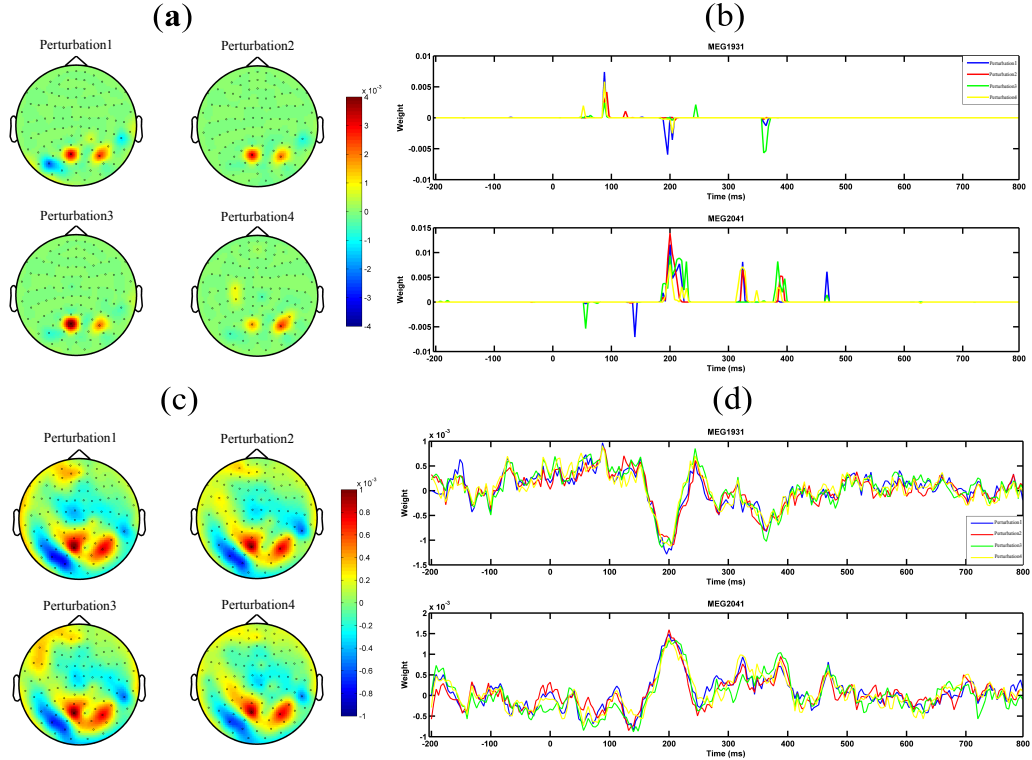}
         \end{subfigure}
         \caption{Comparison of the reproducibility of Lasso when $\delta_\Phi$ and $\zeta_\Phi$ are used in the model selection procedure. (a) and (b) show the spatio-temporal patterns represented by $\vec{\hat{\Theta}}_1^{\delta}$ across the 4 perturbed training sets. (c) and (d) show the spatio-temporal patterns represented by $\vec{\hat{\Theta}}_1^{\zeta}$ across the 4 perturbed training sets. Employing $\zeta_\Phi$ instead of $\delta_\Phi$ in the model selection yields more reproducible MBMs.}
         \label{fig:subj1_Lasso_Stability}
\end{figure}

In addition, optimizing the brain decoding model based on $\zeta_\Phi$ provides more reproducible brain decoders. According to table~\ref{tab:MEG_Sensor_Results}, using $\zeta_\Phi$ instead of $\delta_\Phi$ provides (on average) $0.15$ more reproducibility over 16 subjects. To illustrate the advantage of higher reproducibility on the interpretability of maps, Figure~\ref{fig:subj1_Lasso_Stability} visualizes $\vec{\hat{\Theta}}_1^{\delta}$ and $\vec{\hat{\Theta}}_1^{\zeta}$ over 4 perturbed training sets. The spatial maps [Figure~\ref{fig:subj1_Lasso_Stability}(a) and Figure~\ref{fig:subj1_Lasso_Stability}(c)] are plotted for the magnetometer sensors averaged in the time interval between $184 ms$ and $236 ms$ after stimulus onset. The temporal maps [Figure~\ref{fig:subj1_Lasso_Stability}(b) and Figure~\ref{fig:subj1_Lasso_Stability}(d)] are showing the multivariate temporal maps of MEG1931 and MEG2041 sensors. While $\vec{\hat{\Theta}}_1^{\delta}$ is unstable in time and space across the 4 perturbed training sets, $\vec{\hat{\Theta}}_1^{\zeta}$ provides more reproducible maps.

\section{Discussions}
\label{sec:discussions}
\subsection{Defining Interpretability: Theoretical Advantages}
\label{subsec:theoretical_importance}
An overview of the brain decoding literature shows frequent co-occurrence of the terms interpretation, interpretable, and interpretability with the terms model, classification, parameter, decoding, method, feature, and pattern (see the quick meta-analysis on the literature in the supplementary material); however, a formal formulation of the interpretability is never presented. In this study, our primary interest is to present a theoretical definition of the interpretability of linear brain decoding models and their corresponding MBMs. Furthermore, we show the way in which interpretability is related to the reproducibility and neurophysiological representativeness of MBMs. Our definition and quantification of interpretability remains theoretical, as we assume that the true solution of the brain decoding problem is available. Despite this limitation, we argue that the presented theoretical definition provides a concrete framework of a previously abstract concept and that it establishes a theoretical background to explain an ambiguous phenomenon in the brain decoding context. We support this argument using an example in time-domain MEG decoding in which we show how the presented definition can be exploited to heuristically approximate the interpretability. This example shows how partial prior knowledge\footnote{This partial knowledge can be based on already known facts regarding the timing and location of neural activity.} regarding underlying brain activity can be used to find more plausible multivariate patterns in data. Furthermore, the proposed decomposition of the interpretability of MBMs into their reproducibility and representativeness explains the relationship between the influential cooperative factors in the interpretability of brain decoding models and highlights the possibility of indirect and partial evaluation of interpretability by measuring these effective factors.

\subsection{Application in Model Evaluation}
\label{subsec:model_evaluation}
Discriminative models in the framework of brain decoding provide higher sensitivity and specificity than univariate analysis in hypothesis testing of neuroimaging data. Although multivariate hypothesis testing is performed based solely on the generalization performance of classifiers, the emergent need for extracting reliable complementary information regarding the underlying neuronal activity motivated a considerable amount of research on improving and assessing the interpretability of classifiers and their associated MBMs. Despite ubiquitous use, the generalization performance of classifiers is not a reliable criterion for assessing the interpretability of brain decoding models~\cite{rasmussen2012model}. Therefore, considering extra criteria might be required. However, because of the lack of a formal definition for interpretability, different characteristics of brain decoding models are considered as the main objective in improving their interpretability. Reproducibility~\cite{rasmussen2012model,conroy2013fast}, stability selection~\cite{van2009interpreting,varoquaux2012small,wang2015randomized}, and neurophysiological plausibility~\cite{afshin2011mutual} are examples of related criteria.

Our definition of interpretability helped us to fill this gap by introducing a new multi-objective model selection criterion as a weighted compromise between interpretability and generalization performance of linear models. Our experimental results on single-subject decoding showed that adopting the new criterion for optimizing the hyper-parameters of brain decoding models is an important step toward reliable visualization of learned models from neuroimaging data. It is not the first time in the neuroimaging context that a new metric is proposed in combination with generalization performance for the model selection. Several recent studies proposed the combination of the reproducibility of the maps~\cite{rasmussen2012model,conroy2013fast,strother2014stability} or the stability of the classifiers~\cite{yu2013stability,lim2015estimation} with the performance of discriminative models to enhance the interpretability of decoding models. Our definition of interpretability supports the claim that the reproducibility is not the only effective factor in interpretability. Therefore, our contribution can be considered a complementary effort with respect to the state of the art of improving the interpretability of brain decoding at the model selection level.

Furthermore, this work presents an effective approach for evaluating the quality of different regularization strategies for improving the interpretability of MBMs. As briefly reviewed in Section~\ref{sec:introduction}, there is a trend in research within the brain decoding context in which prior knowledge is injected into the penalization term as a technique to improve the interpretability of decoding models. Thus far, in the literature, there is no ad-hoc method to compare these different methods. Our findings provide a further step toward direct evaluation of interpretability of the currently proposed penalization strategies. This evaluation can highlight the advantages and disadvantages of applying different strategies on different data types and facilitates the choice of appropriate methods for a certain application.

\subsection{Regularization and Interpretability}
\label{subsec:regularization_interpretability}
\citet{haufe2013interpretation} demonstrated that the weight in linear discriminative models does not provide an accurate measure for evaluating the relationship between variables, primarily because of the contribution of noise in the decoding process. This disadvantage is primarily caused by the decoding process that minimizes the classification error only considering the uncertainty in the output space~\cite{aggarwal2009survey,zhang2005support,tzelepis2015linear} and not the uncertainty in the input space (or noise). The authors concluded that the interpretability of brain decoding cannot be improved using regularization. Our experimental results on the toy data (see Section~\ref{subsec:simulation_study}) shows that if the right criterion is used for selecting the best values for hyper-parameters, appropriate choice of the regularization strategy can still play significant role in improving the interpretability of results. For example, in this case, the true generative function behind the sampled data is sparse (see Section~\ref{subsubsec:toy_data}), but because of the noise in the data, the sparse model is not the most accurate one. Using a more comprehensive criterion (in this case, $\zeta_\Phi$) shows the advantage of selecting correct prior assumptions about the distribution of the data via regularization. This observation encourages the modification of the conclusion in~\cite{haufe2013interpretation} as follows: if the performance of the model is the only criterion in the model selection, then the interpretability cannot necessarily be improved by means of regularization.

\subsection{Advantage over Mass-Univariate Analysis}
\label{subsec:univariate}
Mass-univariate hypothesis testing methods are among the most popular tools in neuroscience research because they provide significance checks and a fair level of interpretability via univariate brain maps. Mass-univariate analyses consist of univariate statistical tests on single independent variables followed by multiple comparison correction. Generally, multiple comparison correction reduces the sensitivity of mass-univariate approaches because of the large number of univariate tests involved. Cluster-based permutation testing~\cite{maris2007nonparametric} provides a more sensitive univariate analysis framework by making the cluster assumption in the multiple comparison correction. Unfortunately, this method is not able to detect narrow spatio-temporal effects in the data~\cite{groppe2011amass}. As a remedy, brain decoding provides a very sensitive tool for hypothesis testing; it has the ability to detect multivariate patterns, but suffers from a low level of interpretability. Our study proposes a possible solution for the interpretability problem of classifiers, and therefore, it facilitates the application of brain decoding in the analysis of neuroimaging data. Our experimental results for the MEG data demonstrate that, although the non-parametric cluster-based permutation test is unable to detect the N170 effect in MEG data, employing $\zeta_\Phi$ instead of $\delta_\Phi$ in model selection not only detects the stimuli-relevant information in the data, but also provides both reproducible and representative spatio-temporal mapping of the timing and the location of underlying neurophysiological effect.

\subsection{Limitations and Future Directions}
\label{subsec:limitations_future}
Despite theoretical and practical advantages, the proposed definition and quantification of interpretability suffer from some limitations. All of the theoretical and practical concepts are defined for linear models, with the main assumption that $\Phi^* \in \mathcal{H}$ (where $\mathcal{H}$ is a class of linear functions). This fact highlights the importance of linearizing the experimental protocol in the data collection phase~\cite{naselaris2011encoding}. Extending the definition of interpretability to non-linear models demands future research into the visualization of non-linear models in the form of brain maps. Currently, our findings cannot be directly applied to non-linear models. Furthermore, the proposed heuristic for the time-domain MEG data applies only to binary classification. One possible solution in multiclass classification is to separate the decoding problem into several binary sub-problems. In addition the quality of the proposed heuristic is limited for the small sample size datasets (see supplementary material). Finding physiologically relevant heuristics for other acquisition modalities such as fMRI can be also considered in future work.
\section{Conclusions}
\label{sec:conclusions}
In this paper, we presented a novel theoretical definition for the interpretability of brain decoding and associated multivariate brain maps. We showed how the interpretability relates to the representativeness and reproducibility of brain decoding. The multiplicative nature of the relation between the reproducibility and the representativeness in the computation of interpretability of MBMs is also demonstrated. Although it is theoretical, the presented definition is a first step toward practical solutions for revealing the knowledge learned from linear classifiers. As an example of this major breakthrough, and to provide a proof of concept, a heuristic approach based on the contrast event-related field is proposed for practical evaluation of the interpretability in time-domain MEG decoding. We experimentally showed that adding the interpretability of brain decoding models as a criterion in the model selection procedure yields significantly higher interpretable models by sacrificing a negligible amount of performance in the single-subject decoding scenario. Our methodological and experimental achievements can be considered a complementary theoretical and practical effort that contributes to efforts to enhance the interpretability of multivariate approaches.

\section*{Acknowledgments}
\label{sec:acknowledgments}
The author wishes to thank Sandro Vega-Pons and Nathan Weisz for valuable discussions and comments.

\appendix 
\section{cERF and its Generative Nature}
\label{sec:cERF_generative}
According to~\cite{haufe2013interpretation}, for a linear discriminative model with parameters $\Theta$, the unique equivalent generative model can be computed as follows:
\begin{eqnarray}
\label{eq:haufe}
A \propto \Sigma_\textbf{X} \Theta
\end{eqnarray}
In a binary ($\textbf{Y}=\{1,-1\}$) least squares classification scenario, we have:
\begin{eqnarray}
\label{eq:haufe}
A \propto \Sigma_\textbf{X} \Sigma_\textbf{X}^{-1} \textbf{X}^T \textbf{Y} =  \textbf{X}^T \textbf{Y} = \mu^+ - \mu^-
\end{eqnarray}
where $\Sigma_\textbf{X}$ represents the covariance of the input matrix $\textbf{X}$, and $\mu^+$ and $\mu^-$ are the means of positive and negative samples, respectively. Therefore, the equivalent generative model for the above classification problem can be derived by computing the difference between the mean of samples in two classes, which is equivalent to the definition of cERF in time-domain MEG data.

\section{Relation between $\beta_\Phi$ and $\tilde{\beta}_\Phi$(Eq. ~\ref{eq:representativeness_estimation2})}
\label{sec:missrepresentation}
\begin{figure}
         \centering
		\begin{subfigure}[h!]{0.45\textwidth}
                \includegraphics[width=\textwidth]{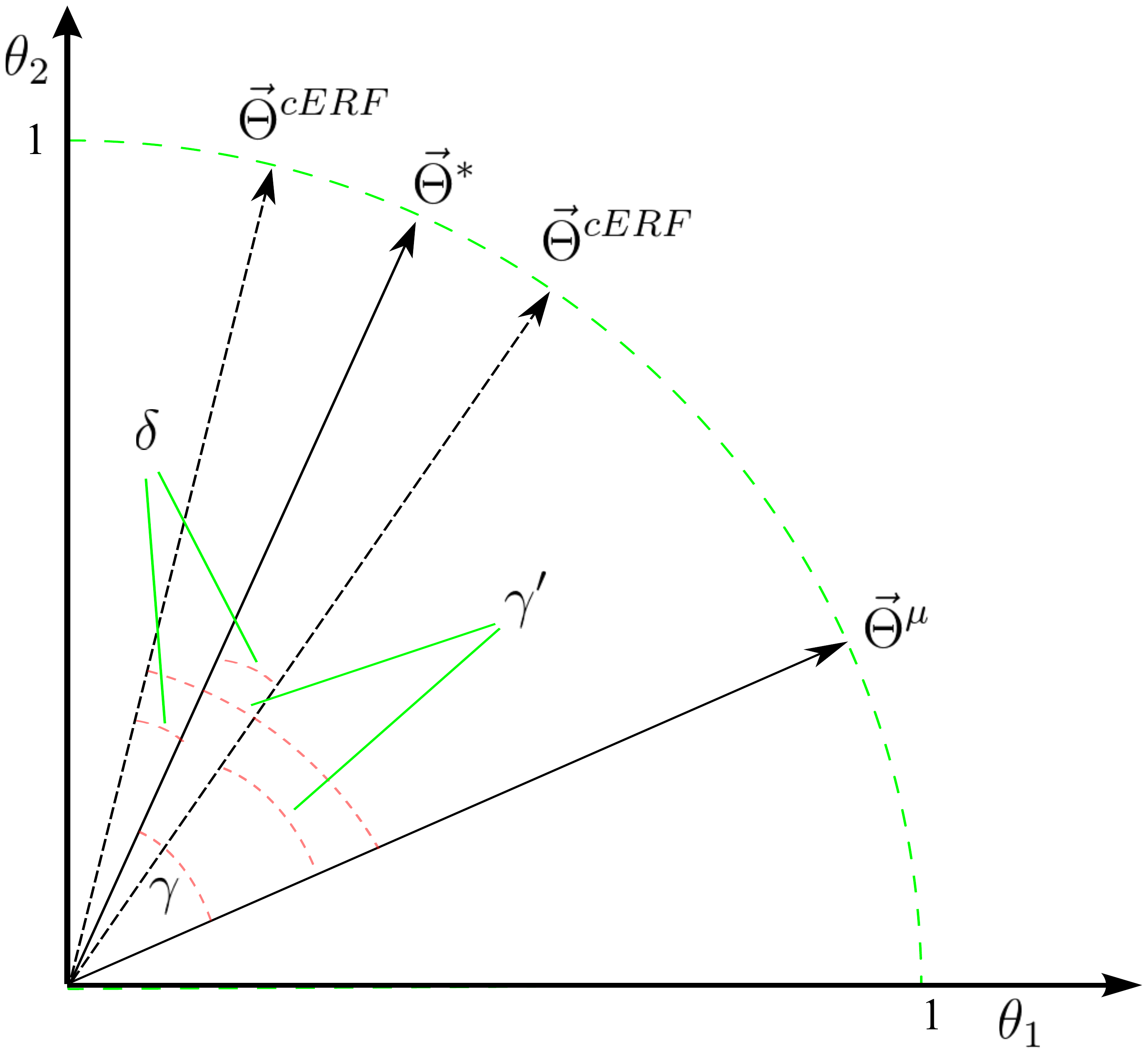}
           		\caption{}
         		\label{subfig:misrepresentativeness}
        \end{subfigure}
        \caption {Misrepresentation of $\vec{\Theta}^{cERF}$ with respect to $\vec{\Theta}^{*}$.}
        \label{fig:misrepresentativeness}
\end{figure}

Let $\gamma$ be the angle between $\vec{\Theta}^\mu$ and $\vec{\Theta}^*$. Let $\gamma'$ be the angle between $\vec{\Theta}^\mu$ and $\vec{\Theta}^{cERF}$. Furthermore, assume that $\delta$ is the angle between $\vec{\Theta}^{*}$ and $\vec{\Theta}^{cERF}$ and that $\Delta_{\beta}=\cos(\delta)$. We consider both cases in which $\beta_\Phi$ is underestimated/overestimated by $\tilde{\beta}_\Phi$ (see Figure~\ref{fig:misrepresentativeness} as an example in 2-dimensional space). Then, we have:

\begin{eqnarray}
\begin{split}
\gamma & = \gamma'\pm\delta \Rightarrow \cos(\gamma) = \cos(\gamma'\pm\delta) \\
& =\cos(\gamma)\cos(\delta)\pm\sin(\gamma)\sin(\delta)=\tilde{\beta}_{\Phi}\Delta_{\beta}\pm\sqrt{(1-\tilde{\beta}^2)(1-\Delta_{\beta}^2)}
\end{split}
\end{eqnarray}

\section{Relation between $\eta_\Phi$ and $\tilde{\eta}_\Phi$(Eq. ~\ref{eq:interpretability_estimation2})}
\label{sec:interpretation_estimation}
\begin{figure}
         \centering
		\begin{subfigure}[h!]{0.45\textwidth}
                \includegraphics[width=\textwidth]{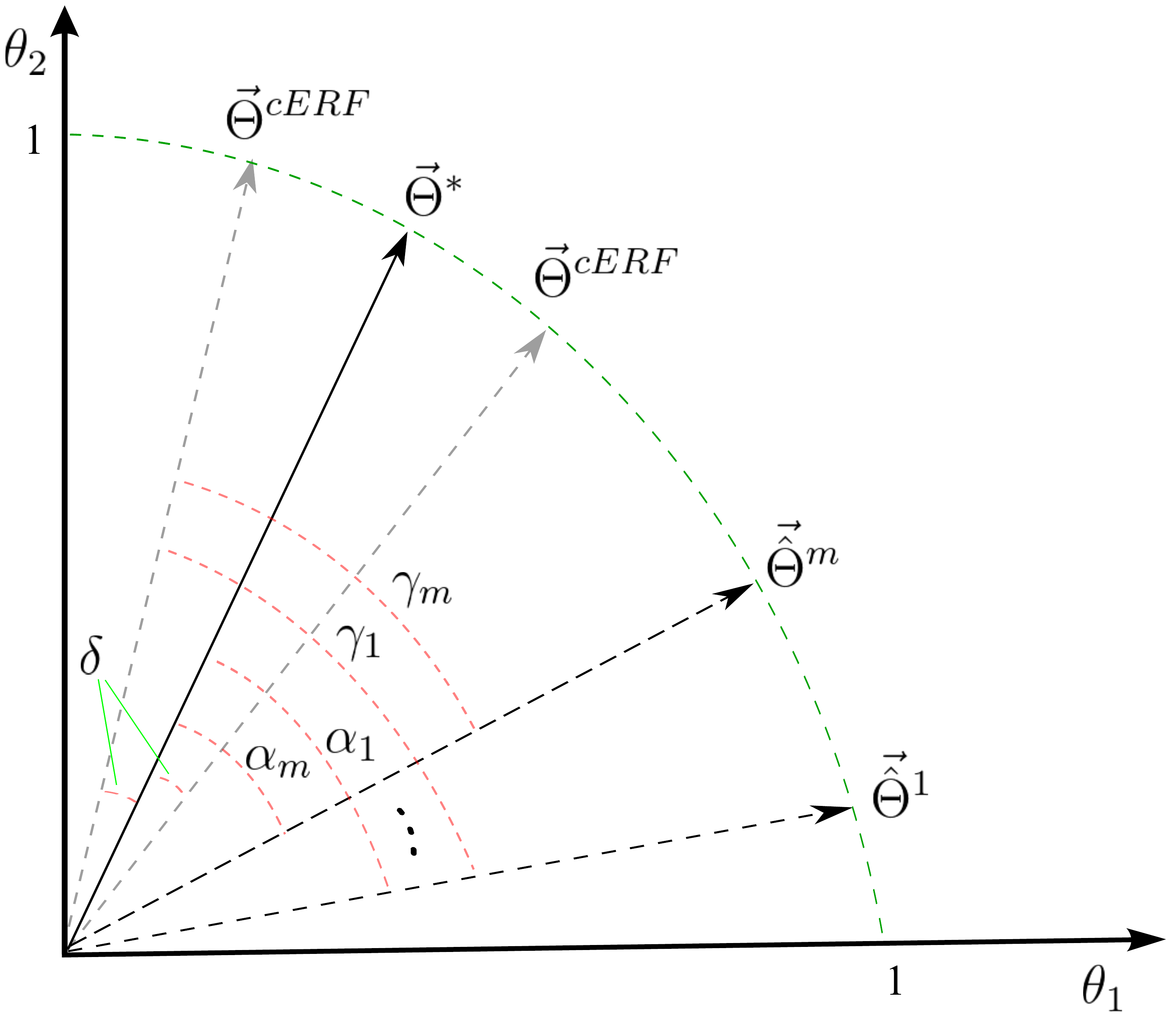}
           		\caption{}
         		\label{subfig:interpretability_estimation}
        \end{subfigure}
        \caption {Relation between $\eta_\Phi$ and $\tilde{\eta}_\Phi$.}
        \label{fig:interpretability_estimation}
\end{figure}

Let $\alpha_1, \dots, \alpha_m$ be the angles between $\vec{\hat{\Theta}}^1, \dots, \vec{\hat{\Theta}}^m$ and $\vec{\Theta}^{*}$, and $\gamma_1, \dots, \gamma_m$ be the angles between $\vec{\hat{\Theta}}^1, \dots, \vec{\hat{\Theta}}^m$ and $\vec{\Theta}^{cERF}$. Furthermore, assume that $\delta$ is the angle between $\vec{\Theta}^{*}$ and $\vec{\Theta}^{cERF}$. We consider both cases in which $\eta_\Phi$ is underestimated/overestimated by $\tilde{\eta}_\Phi$ (see Figure~\ref{fig:interpretability_estimation} as an example in 2-dimensional space).

\begin{eqnarray}
\begin{split}
\eta_\Phi & = \frac{\cos(\alpha_1)+ \dots + \cos(\alpha_m)}{m}
= \frac{\cos(\gamma_1\pm\delta)+ \dots + \cos(\gamma_m\pm\delta)}{m} \\
& = \frac{\cos(\gamma_1)\cos(\delta)\pm\sin(\gamma_1)\sin(\delta)+ \dots +\cos(\gamma_m)\cos(\delta)\pm\sin(\gamma_m)\sin(\delta)}{m} \\
& \xrightarrow{\Delta_{\beta}=\cos(\delta)}
= \frac{\Delta_{\beta}[\cos(\gamma_1)+\dots+\cos(\gamma_m)] \pm \sin(\delta)[\sin(\gamma_1)+\dots+\sin(\gamma_m)]}{m} \\
& \xrightarrow{\tilde{\eta}_\Phi=\frac{\cos(\gamma_1)+\dots+\cos(\gamma_m)}{m}}
\eta_\Phi = \Delta_{\beta}\tilde{\eta}_\Phi \pm \frac{\sqrt{1-\Delta_{\beta}^2}}{m}(\sin(\gamma_1)+\dots+\sin(\gamma_m))
\end{split}
\end{eqnarray}

\section{Proof of Proposition~\ref{theo:interpretability}}
\label{sec:interpretability_proof}
\begin{figure}
         \centering
		 \begin{subfigure}[h!]{0.45\textwidth}
                \includegraphics[width=\textwidth]{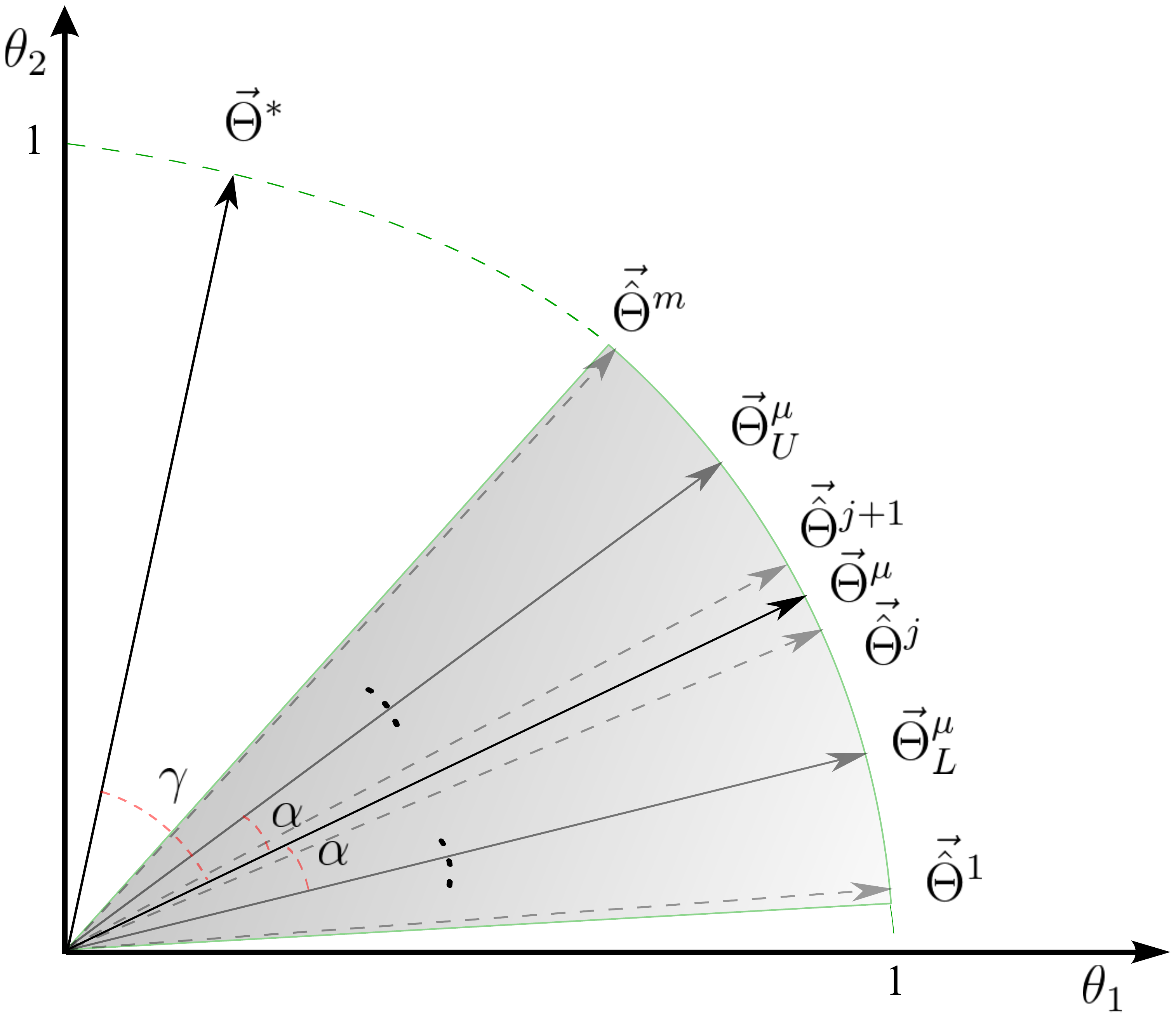}
          		\caption{}
         		\label{subfig:interpretability_theory}
        \end{subfigure}
        \caption {Relation between representativeness, reproducibility, and interpretability in 2 dimensions.}
        \label{fig:interpretability_theory}
\end{figure}

Throughout this proof, we assume that all of the parameter vectors are normalized in the unit hypersphere (see Figure~\ref{fig:interpretability_theory} as an illustrative example in 2 dimensions). Let $T = \{\vec{\hat{\Theta}}^1, \dots ,\vec{\hat{\Theta}}^m\}$ be a set $m$ MBMs, for $m$ perturbed training sets where $\vec{\hat{\Theta}}^i \in \mathbb{R}^p$. Now, consider any arbitrary $p-1$-dimensional hyperplane $\mathcal{A}$ that contains $\vec{\Theta}^\mu$. Clearly, $\mathcal{A}$ divides the $p$-dimensional parameter space into 2 subspaces. Let $\triangledown$ and $\blacktriangledown$ be binary operators where $\vec{\Theta}^i \triangledown \vec{\Theta}^k$ indicates that $\vec{\Theta}^i$ and $\vec{\Theta}^k$ are in the same subspace, and $\vec{\Theta}^i \blacktriangledown \vec{\Theta}^k$ indicates that they are in different subspaces. Now, we define $T_U=\{\vec{\Theta}^i \mid \vec{\Theta}^i \triangledown \vec{\Theta}^*\}$ and $T_L=\{\vec{\Theta}^i \mid \vec{\Theta}^i \blacktriangledown \vec{\Theta}^*\}$. Let the cardinality of $T_L$ denoted by $n(T_L)$ be $j$ ($n(T_L)=j$). Thus, $n(T_U)=m-j$. Now, assume that $\measuredangle(\vec{\hat{\Theta}}^i,\mathcal{A}) = \alpha_1,\dots,\alpha_j$ are the angles between $\vec{\hat{\Theta}}^i \in T_L$ and $\mathcal{A}$, and (similarly) $\alpha_{j+1},\dots,\alpha_m$ for $\vec{\hat{\Theta}}^i \in T_U$ and $\mathcal{A}$. Based on Eq.~\ref{eq:main_map}, let $\vec{\Theta}_L^\mu$ and $\vec{\Theta}_U^\mu$ be the main maps of $T_L$ and $T_U$, respectively. Therefore, we obtain $\vec{\Theta}^\mu = \frac{\vec{\Theta}_L^\mu+\vec{\Theta}_U^\mu}{\left \|\vec{\Theta}_L^\mu+\vec{\Theta}_U^\mu \right \|}$ and $\measuredangle(\vec{\Theta}_L^\mu,\mathcal{A})= \measuredangle(\vec{\Theta}_U^\mu,\mathcal{A})= \alpha$. Furthermore, assume $\measuredangle(\vec{\Theta}^*,\mathcal{A})=\gamma$. As a result, $\psi_\Phi=\cos(\alpha)$ and $\beta_\Phi=\cos(\gamma)$. According to Eq.~\ref{eq:interpretability} and using a cosine similarity definition, we have:

\begin{eqnarray}
\begin{split}
\eta_\Phi & = \frac{1}{m}\sum_{j=1}^{m} \left |\vec{\Theta}^*.\vec{\hat{\Theta}}^j\right | \\
& = \frac{\cos(\gamma+\alpha_1)+ \dots + \cos(\gamma+\alpha_j)+\cos(\gamma-\alpha_{j+1})+ \dots + \cos(\gamma-\alpha_m)}{m} \\
& = \frac{\cos(\gamma+\alpha)+ \cos(\gamma-\alpha)}{2}  \\
& = \frac{\cos(\gamma)\cos(\alpha)-\sin(\gamma)\sin(\alpha)+\cos(\gamma)\cos(\alpha)+\sin(\gamma)\sin(\alpha)}{2} \\
& = \cos(\gamma)\cos(\alpha) = \beta_\Phi \times \psi_\Phi.
\end{split}
\end{eqnarray}

A similar procedure can be used to prove $\tilde{\eta}_\Phi=\tilde{\beta}_\Phi \times \psi_\Phi$  by replacing $\vec{\Theta}^*$ with $\vec{\Theta}^{cERF}$.

\section{Computing the Bias and Variance in Binary Classification}
\label{sec:bias_variance_computation}
Here, using the out-of-bag (OOB) technique, and based on procedures proposed by~\cite{domingos2000unified} and~\cite{valentini2004bias}, we compute the expected prediction error (EPE) for a linear binary classifier $\Phi$ under bootstrap perturbation of the training set. Let $m$ be the number of perturbed training sets resulting from partitioning $(X,Y)$ into $(X_{tr},Y_{tr})$ and $(X_{ts},Y_{ts})$, i.e., training and test sets. If $\hat{\Phi}^j$ is the linear classifier estimated from the $j$th perturbed training set, then the main prediction $\Phi^\mu(\textbf{x}_i)$ for each sample in the dataset can be computed as follows:

\begin{eqnarray} \label{eq:mean_prediction}
\Phi^\mu(\textbf{x}_i)=\left\{\begin{matrix}
1 & if \quad \frac{1}{k_i}\sum_{j=1}^{k_i}\hat{\Phi}^j(\textbf{x}_i)\geq \frac{1}{2}   \\
0 & otherwise
\end{matrix}\right.
\end{eqnarray}

where $k_i$ is the number of times that $x_i$ is present in the test set\footnote{It is expected that each sample $\textbf{x}_i \in X$ appears (on average) $k_i\approx\frac{m}{3}$ times in the test sets.}.1

The computation of bias is challenging because the optimal model $\Phi^*$ is unknown. According to \cite{tibshirani1996bias}, misclassification error is one of the loss measures that satisfies a Pythagorean-type equality, and:
\begin{eqnarray} \label{eq:bias_equality}
\frac{1}{n}\sum_{i=1}^{n}\mathcal{L}(\Phi^\mu(\textbf{x}_i),\Phi^*(\textbf{x}_i))=\frac{1}{n}\sum_{i=1}^{n}\mathcal{L}(y_i,\Phi^\mu(\textbf{x}_i))-
\frac{1}{n}\sum_{i=1}^{n}\mathcal{L}(y_i,\Phi^*(\textbf{x}_i))
\end{eqnarray}

Because all terms of the above equation are positive, the mean loss between the main prediction and the actual labels can be considered as an upper-bound for the bias:
\begin{eqnarray} \label{eq:bias_inequality}
\frac{1}{n}\sum_{i=1}^{n}\mathcal{L}(\Phi^\mu(\textbf{x}_i),\Phi^*(\textbf{x}_i))\leq \frac{1}{n}\sum_{i=1}^{n}\mathcal{L}(y_i,\Phi^\mu(\textbf{x}_i))
\end{eqnarray}

Therefore, a pessimistic approximation of bias $B(\textbf{x}_i)$ can be calculated as follows:
\begin{eqnarray} \label{eq:bias_approximation}
B(\textbf{x}_i)=\left\{\begin{matrix}
0 & if \quad \Phi^\mu(\textbf{x}_i) = y_i \\
1 & otherwise
\end{matrix}\right.
\end{eqnarray}

Then, the unbiased and biased variances (see~\cite{domingos2000unified} for definitions) in each training set can be calculated by:

\begin{eqnarray} \label{eq:unbiased_variance}
V_{u}^j(\textbf{x}_i)=\left\{\begin{matrix}
1 & if \quad B(\textbf{x}_i) = 0 \quad and \quad \Phi^\mu(\textbf{x}_i) \neq \hat{\Phi}^j(\textbf{x}_i) \\
0 & otherwise
\end{matrix}\right.
\end{eqnarray}

\begin{eqnarray} \label{eq:biased_variance}
V_{b}^j(\textbf{x}_i)=\left\{\begin{matrix}
1 & if \quad B(\textbf{x}_i) = 1 \quad and \quad \Phi^\mu(\textbf{x}_i) \neq \hat{\Phi}^j(\textbf{x}_i) \\
0 & otherwise
\end{matrix}\right.
\end{eqnarray}

Then, the expected prediction error of $\Phi$ can be computed as follows (ignoring the irreducible error):

\begin{eqnarray} \label{eq:EPE_Partitioning}
\begin{split}
EPE_{\Phi}(X)=\underbrace{\frac{1}{n}\sum_{i=1}^{n}B(\textbf{x}_i)}_{Bias}+
\\ \underbrace{\frac{1}{nm}\sum_{j=1}^{m}\sum_{i=1}^{n}[V_{u}^j(\textbf{x}_i)-V_{b}^j(\textbf{x}_i)]}_{Variance}
\end{split}
\end{eqnarray}


\bibliographystyle{model1-num-names}
\small
\bibliography{Interpretation}

\end{document}